\documentclass[pdflatex,sn-mathphys-num]{sn-jnl}

\usepackage{algorithm2e}
\usepackage{graphicx}%
\usepackage{multirow}%
\usepackage{amsmath,amssymb,amsfonts}%
\usepackage{amsthm}%
\usepackage{mathrsfs}%
\usepackage[title]{appendix}%
\usepackage{xcolor}%
\usepackage{textcomp}%
\usepackage{manyfoot}%
\usepackage{booktabs}%
\usepackage{algorithmicx}%
\usepackage{algpseudocode}%
\usepackage{listings}%
\usepackage{enumitem}

\theoremstyle{thmstyleone}%
\theoremstyle{thmstyletwo}%

\theoremstyle{thmstylethree}%

\begin{document}

\title[Article Title]{PCLVis: Visual Analytics of Process Communication Latency in Large-Scale Simulation}








\author[1]{\fnm{Chongke} \sur{Bi}}\email{bichongke@tju.edu.cn}

\author[1]{\fnm{Xin} \sur{Gao}}\email{gao\_xin\_private@163.com}

\author[1]{\fnm{Baofeng} \sur{Fu}}\email{fubaofeng96@163.com}

\author[2]{\fnm{Yuheng} \sur{Zhao}}\email{22110980020@fudan.edu.cn}

\author[2]{\fnm{Siming} \sur{Chen}}\email{simingchen@fudan.edu.cn}

\author[3]{\fnm{Ying} \sur{Zhao}}\email{zhaoying@csu.edu.cn}

\author*[4,5]{\fnm{Lu} \sur{Yang}}\email{yanglu8206@163.com}

\affil*[1]{\orgdiv{College of Intelligence and Computing}, \orgname{Tianjin University}, \orgaddress{\city{Tianjin}, \postcode{300072}, \country{China}}}

\affil[2]{\orgdiv{School of Data Science}, \orgname{Fudan University}, \orgaddress{\city{Shanghai}, \postcode{200433}, \country{China}}}

\affil[3]{\orgdiv{School of Computer Science and Engineering}, \orgname{Central South University}, \orgaddress{\city{Changsha}, \postcode{410083}, \country{China}}}

\affil[4]{\orgdiv{Tianjin Key Laboratory of Advanced Electromechanical System Design and Intelligent Control}, \orgname{Tianjin University of Technology}, \orgaddress{\city{Tianjin}, \postcode{300384}, \country{China}}}

\affil[5]{\orgdiv{National Demonstration Center for Experimental Mechanical and Electrical Engineering Education }, \orgname{Tianjin University of Technology}, \orgaddress{\city{Tianjin}, \postcode{300384}, \country{China}}}


\abstract{Large-scale simulations on supercomputers have become important tools for users.
However, their scalability remains a problem due to the huge communication cost among parallel processes.
Most of the existing communication latency analysis methods rely on the physical link layer information, which is only available to administrators.
In this paper, a framework called PCLVis is proposed to help general users analyze process communication latency (PCL) events.
Instead of the physical link layer information, the PCLVis uses the MPI process communication data for the analysis.
First, a spatial PCL event locating method is developed.
All processes with high correlation are classified into a single cluster by constructing a process-correlation tree.
Second, the propagation path of PCL events is analyzed by constructing a communication-dependency-based directed acyclic graph (DAG), which can help users interactively explore a PCL event from the temporal evolution of a located PCL events cluster.
In this graph, a sliding window algorithm is designed to generate the PCL events abstraction. Meanwhile, a new glyph called communication state glyph (CS-Glyph) is designed for each process to show its communication states, including its I/o messages and load balance.
Each leaf node can be further unfolded to view additional information.
Third, a PCL event attribution strategy is formulated to help users optimize their simulations.
The effectiveness of the PCLVis framework is demonstrated by analyzing the PCL events of several simulations running on the TH-1A supercomputer.
By using the proposed framework, users can greatly improve the efficiency of their simulations.}




\maketitle

\section{Introduction}\label{sec:introduction}

Large-scale simulations have found widespread utility across diverse fields, including fluid mechanics, aerodynamics, and aerospace. Leveraging the computational might of supercomputers for virtual simulations offers a promising means to conserve precious resources. These supercomputer systems comprise two foundational components: compute nodes and communication network infrastructure. Each computing node is equipped with its distinct private memory space, ensuring isolation from other nodes. In the context of large-scale simulations, supercomputers partition computational tasks into discrete processes for parallel execution across distinct compute nodes. These processes necessitate communication for synchronized operation and data exchange, highlighting the centrality of process communication.

\begin{figure*}[!htb]
\centering
\includegraphics[width=0.85\textwidth]{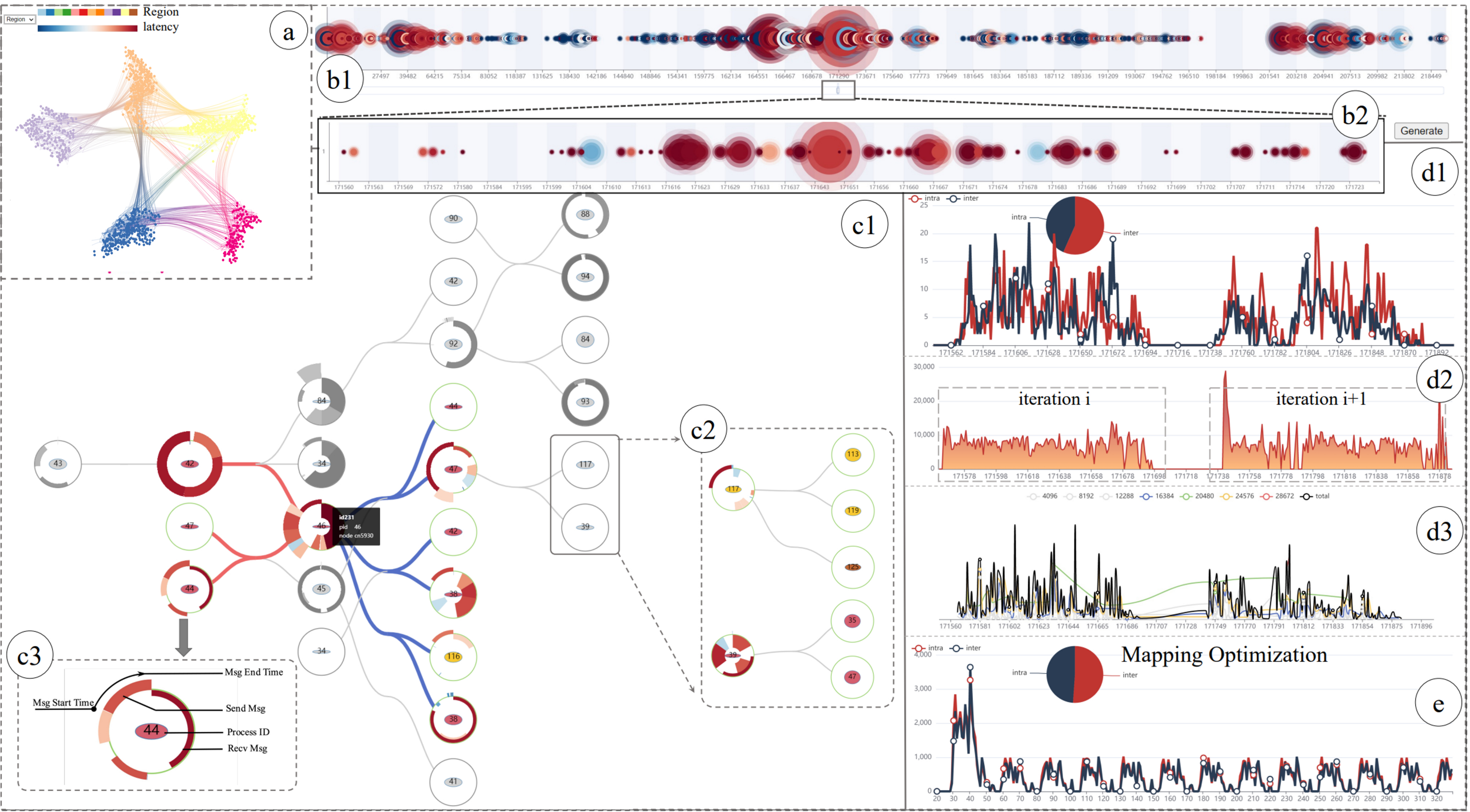}
\caption{Visual analytics of the PCL events from a parallel application running with 1152 processes on a TH-1A supercomputer.
(a) shows the spatio-PCL event clustering results obtained using our process-correlation-tree-based algorithm. The temporal evolution of each cluster can be viewed in (b1), and the details can be further interactively explored in (b2).
(c) shows the constructed communication-dependency-based DAG graph, which helps users further explore the extracted PCL events. For each selected process, its former and latter connected processes are denoted by the red and blue curves, respectively.
The leaf process can be further unfolded as shown in (c2), and the design of each glyph is shown in (c3).
The possible PCL causes as summarized in (d) include (d1) poor process-to-processor mapping (too much intercommunication), (d2) poor communication pattern (unbalanced workload), and (d3) background traffic due to the application of other users.
(e) shows the result obtained using an optimized process-to-processor mapping. The ratio between the intra- and inter-communications became larger than the original value reported in (d1).}
\label{fig:teaser}
\end{figure*}

However, the performance of simulation applications can be curtailed by bottlenecks in process communication\cite{Martinez:2015}. Communication delay events can unpredictably manifest at any juncture, prompting an imperative to dissect the underlying causes. Existing methods for communication latency analysis predominantly lean on the physical link layer, furnishing granular information encompassing start and end points, routing paths, and intermediate waypoints (routers). This information empowers the accurate identification of communication latency events. Nonetheless, the accessibility of physical link layer insights is confined to system administrators, limiting general users' access.

In the absence of access to physical link layer data, general users are relegated to investigating communication delays via Message Passing Interface (MPI) process communication data. To this end, we initiate our analysis by preprocessing the data. Specifically, we curate a communication event dataset through the scrutiny of parallel execution traces recorded by the Tuning Analysis Utility (TAU), yielding essential communication trace data. However, while MPI process communication data encapsulates the communication's origin (Source MPI rank) and destination (destination MPI rank), the absence of routing path and router details poses two fundamental challenges~\cite{Kuhn:2014}. Firstly, pinpointing the latency region becomes intricate, given that delays often manifest at intermediary waypoints such as routers or routing paths. Secondly, unraveling the propagation path of communication delay events proves complex. The propagation path, integral to the routing path, remains elusive to general users devoid of access to the physical link layer.

To bridge this gap, we introduce 'PCLVis', an innovative framework tailored for the wider supercomputer user base. PCLVis facilitates the visual analysis of process communication latency (PCL) events utilizing readily available MPI process communication data.
First, a space-PCL event localization method is developed to localize processes involved in PCL events.
All processes with a high correlation are classified into a single cluster by constructing a process-correlation tree.
Second, the propagation path of PCL events is analyzed by constructing a communication-dependency-based directed acyclic graph (DAG), which can help users interactively explore a PCL event from the temporal evolution of a located PCL events cluster.
In the end, A PCL events attribution strategy is also designed to help users optimize their simulations.
And in summary, the graphics user interface of PCLVis is shown in Figure~\ref{fig:teaser}.

Based on the above works, PCLVis is evaluated via visual analytics of three types of communication data on a supercomputer.
Results show that the proposed framework can help users greatly improve the efficiency of their simulations.

The main contributions of this work include:
\begin{itemize}
\item A process-correlation-tree-based spatio-PCL event locating method is proposed to help users locate those communication regions with high latency.
\item A communication-dependency-based DAG is constructed to help users track the propagation path of PCL events.
\item A PCL events attribution strategy is designed to help users optimize their simulations.

\end{itemize}

\section{Related Work}\label{sec:related}
This section reviews some existing work on communication delay analysis methods (Section~\ref{ssec:related-delay}), processes communication visualization methods (Section~\ref{ssec:related-visualization}),
and communication delay attribution methods (Section~\ref{ssec:related-attribution}).

\subsection{Communication Delay Analysis}\label{ssec:related-delay}
Analyzing communication delays from a large-scale simulation is a challenging task due to the large data size.
Some studies have attempted to reduce the data size by using clustering~\cite{Nickolayev:1997,Aguilera:2006,Gamblin:2010,zan2022high}, compression~\cite{Knupfer:2003,Lee:2008,Noeth:2009,Zhai:2014}, and some other algorithms~\cite{Xu:2009,Wu:2013}.
Nevertheless, the precision of analysis results does not satisfy the user requirements because some important features have been lost during the data reduction process.
Several analysis tools can visualize the basic and statistical information of communication events, like the scalasca performance toolset~\cite{Geimer:2010},
multi-threaded parallel application analysis tool~\cite{De:2000},
the TAU parallel performance tool~\cite{Shende:2006},
the HPCToolkit~\cite{Adhianto:2010},
the Craypat-cray X1 performance analysis tool~\cite{Kaufmann:2003},
and some other tools~\cite{Mohr:2003,Wilke:2020,Geimer:2009}.
However, they did not analyze communication delay events.
Some researchers have proposed structure-based analysis methods. For instance, Hendriks et al.~\cite{Hendriks:2012,Hendriks:2017} proposed critical-path analysis and distance analysis algorithms to help users track communication delays.
Execution phases have also been used to track the communication state in different running phases~\cite{Casas:2007,Chetsa:2013,Gonzalez:2014}.
Pattern matching methods~\cite{Kunz:1997,Kockerbauer:2010,Wolf:2007} have been proposed to detect predefined communication patterns and have successfully extracted inefficient communication behaviors.
Isaacs et al.~\cite{Isaacs:2014-Extracting,Isaacs:2015-Recovering,Isaacs:2015-Ordering}
proposed a series of methods that can help users detect communication delays by extracting a logical structure from parallel tracking data.
While these methods can help users detect communication delays, their accuracy cannot be guaranteed.
To address this problem, this paper proposes a spatial-temporal communication delay locating method that can successfully locate all communication delays from large-scale communication data, thereby helping users further analyze these delays.

\subsection{Process Communication Visualization}\label{ssec:related-visualization}
Process communication is a type of event sequence data. Therefore, this section not only examines process communication but also surveys the visualization of other event sequences.
Several visualization tools have been developed, including VAMPIR~\cite{Nagel:1996},
Paragraph~\cite{Heath:1994}, Paraver~\cite{Pillet:1995}, Projections~\cite{Kale:2003}, which can visualize the performance of a simulation using a Gantt chart.
However, these tools cannot support users in further exploring the details of process communication.
They also cannot be directly used to analyze large-scale simulations due to the limitations in their scalability.
Some improvement methods~\cite{Huang:2009,Jo:2014,Han:2015} have been proposed by making good use of sorting algorithms.
While these Gantt-chart-based methods can be used to visualize event sequences, they still do not fully meet user requirements, especially for visualizing long event sequence data, including process communication.

To visualize a large-scale event sequence, flow-based methods can offer a visualization abstract for users~\cite{Wongsuphasawat:2012} by aggregating event sequences~\cite{Krstajic:2011}.
Some scholars have attempted to reveal the sequence evolution pattern by using aggregated visualization methods, including simplification~\cite{hou2021mswap}, flow design~\cite{Gotz:2014,wang2022ultralight}, feature extraction~\cite{Du:2016}, progression analysis~\cite{Guo:2018}, and some methods for a specific application~\cite{Xu:2016,Mu:2019}. 
When visualizing a large-scale communication event sequence, scale-related issues are addressed by changing the vertical axis from processes to event duration~\cite{Muelder:2009,Sigovan:2013}.
Isaacs et al.~\cite{Isaacs:2014-Combing} reduced visual clutter by adopting a layered abstraction algorithm to visualize the logicalized process communication sequence.
LBVis~\cite{Zhang:2020} applies an interactive visualization method to show the data transmission among different processes.
While these methods can successfully visualize large-scale communication data, they do not offer communication dependency and state information to users, which are vital for understanding the evolution of communication latency.
To address these limitations, this paper constructs a communication-dependency-based DAG that can help users track communication latency.
A CS-Glyph is also designed to show the detailed state of a process.

\subsection{Communication Delay Attribution}\label{ssec:related-attribution}
Communication delay attribution is important for users to improve the efficiency of their simulation.
Most existing communication delay attribution methods mainly rely on analyzing the link layer log information~\cite{Landge:2012,Bhatele:2012,wang2021predicting,Bhatele:2016}.
In other words, these methods utilize the logs in routers through which messages have passed to check for possible communication latency.
Fujiwara et al.~\cite{Fujiwara:2017} evaluated the transmission efficiency in a communication path using hop bytes and designed a re-routing and remapping algorithm to optimize the communication efficiency of a simulation.
Li et al.~\cite{Li:2017-Visual} analyzed the performance of simulations in different workloads and routing strategies by using physical link layer information.
Jha et al.~\cite{Jha:2020} summarized the causes of communication latency by detecting network congestion using link layer information.
Meanwhile, Taffet et al.~\cite{Taffet:2019} attributed communication delays to three causes, namely, poor process-to-processor mapping (placement of MPI ranks on physical cores), poor communication patterns, and background traffic due to the application of other users.

A communication pattern illustrates how processes send and receive data to one another.
Some example patterns include the many-to-one and nearest-neighbor communication patterns.
All of these patterns belong to an unbalanced workload~\cite{Haghi:2021}.
A poor communication pattern can lead to congestion and communication latency.
Demmel et al.~\cite{Demmel:2012} reduced latency events using the communication avoidance algorithm.
A poor process-to-processor mapping only extends the message transmission path, thereby increasing communication latency.
Yan et al.~\cite{YanBaicheng:2019} greatly improved communication performance by designing a good process-to-processor mapping.
Background traffic is considered a job interference~\cite{Yang:2016}.
The network of a supercomputer is shared by many users, thereby restricting the performance of their simulations.

The majority of the above methods rely on link layer log information, which is only available to the administrators of a supercomputer.
Therefore, this paper designs PCLVis as a process-information-based communication delay attribution method to help users analyze the causes of communication latency and improve the efficiency of their simulations.

\section{OVERVIEW OF PCLVis}\label{sec:overview}

This section summarizes the user requirements (Section~\ref{ssec:requirements}) and presents the PCLVis design (Section~\ref{ssec:overview}).

\subsection{User Requirements}\label{ssec:requirements}
The requirements presented in this section were gathered through comprehensive interviews with a diverse group of supercomputer users. With insights derived from these interviews, we have distilled the users' needs into the following synthesized set of requirements:

\textbf{R}1: Locate high latency region.
In large-scale supercomputer simulations, users face a complex communication environment. 
They need an intuitive way to understand communication dynamics throughout the simulation, enabling comprehensive analysis of design mechanisms. 
However, the sheer volume of communication data exchanged among processes presents a challenge. 
Variability in latency due to supercomputer background traffic adds complexity, making it impractical to set a fixed latency threshold. 
To address these issues, our system efficiently locates all latency instances in extensive communication data.

\textbf{R}2: Explore temporal evolution of latency.
In extensive, long-duration simulations spanning days or weeks, users aim to understand how communication delays evolve over time in specific regions. 
This understanding helps identify and grasp periods of increased latency, enabling timely adjustments and effective responses during the entire simulation. 
Our system is designed to facilitate the extraction of crucial latency periods, empowering users to delve into comprehensive analyses of communication latencies over time.

\textbf{R}3: Communication delay attribution.
Users require an efficient and user-friendly approach to assess communication latencies and their origins. 
This is vital for enabling ordinary users to engage with the simulation effectively. 
Our system meets this need by clearly depicting communication latency dependencies and summarizing their potential sources, making it easier for users to comprehend logical connections and identify underlying causes.

\textbf{R}4: Offer possible optimization schemes.
Following the analysis of delay causes, users seek actionable strategies to counter communication delays within diverse timeframes and regions. 
They anticipate receiving initial optimization approaches or guidance to initiate effective improvements. 
In response to this, our system is designed to offer a range of potential optimization schemes, empowering users to proactively address communication delays through informed decisions.

\subsection{System Overview}\label{ssec:overview}
According to the abovementioned requirements, we design PCLVis for a visual analysis of PCL events in a large-scale simulation. 

To analyze communication delay using MPI process communication data, we first preprocess the data. Specifically, we built a communication event dataset by analyzing parallel execution traces collected by the Tuning Analysis Utility (TAU) to obtain communication trace data~\cite{Shende:2006}.
During the runtime of the simulation program, TAU intelligently detects functions, methods, and code blocks, thereby capturing essential communication tracing data.
The properties of a communication event are as follows:

\begin{itemize}
\setlength{\itemsep}{0pt}
\setlength{\parsep}{0pt}
\setlength{\parskip}{0pt}
\item \emph{MPI Rank}: Unique process to which the event belongs.
\item \emph{Type}: Two types of communication events (i.e., send or receive).
\item \emph{Timestamp}: Wall clock time of the event.
\item \emph{Source}: Source MPI rank of the event.
\item \emph{Destination}: Destination MPI rank of the event.
\item \emph{Message Size}: Size of message within the event.
\end{itemize}

PCLVis has three main parts, namely, spatial latency, temporal latency, and attribution. First, the spatial latency part offers information about communication latency within regions. In this part, a spatio-PCL event locating method is proposed to help users locate those communication regions with high latency (\textbf{R1}). Second, the temporal latency of these communication regions is analyzed by DAG. We offer an evolution view to display the latency abstraction over time. Users can select a period of high latency to further explore the communication latency among processes in DAG (\textbf{R2}). Third, the attribution part offers three views to display information about the process of communication states. Users can analyze the causes of communication latencies using our proposed attribution strategy (\textbf{R3}) and then optimize their simulations using an appropriate method that is provided through our system. Such methods may include optimizing the physical mapping of nodes, adjusting the communication patterns, and choosing the running time with better background traffic (\textbf{R4}).

\section{LOCATING SPATIAL PROCESS COMMUNICATION LATENCY EVENTS}\label{sec:locating}
This section introduces our spatio-PCL events locating method (\textbf{R1}).
A process-correlation tree is constructed to divide all processes into different clusters (Section~\ref{ssec:spatial-clustering}), and several criteria for defining a PCL event and the latency of a communication region are prepared (Section~\ref{ssec:criteria}).

\subsection{Criteria for Defining Communication Latency Within a Region}\label{ssec:criteria}

Given that a supercomputer is shared among users, the theoretical transmission speed cannot be used to define a PCL event.
At the same time, there is no proper way to describe a communication delay standard in the given region.
Therefore, this paper designs a statistical method for defining a PCL event in a region.

In the following tasks, we define two latency criteria (transmission time) for intra-node and inter-node communication.
The difference between these criteria lies in whether two communication processes are in the same physical nodes or not given that the transmission speed of inter-node communication is much lower than that of intra-node communication.

\begin{figure}[tb]
\centering
\includegraphics[width=3.4in]{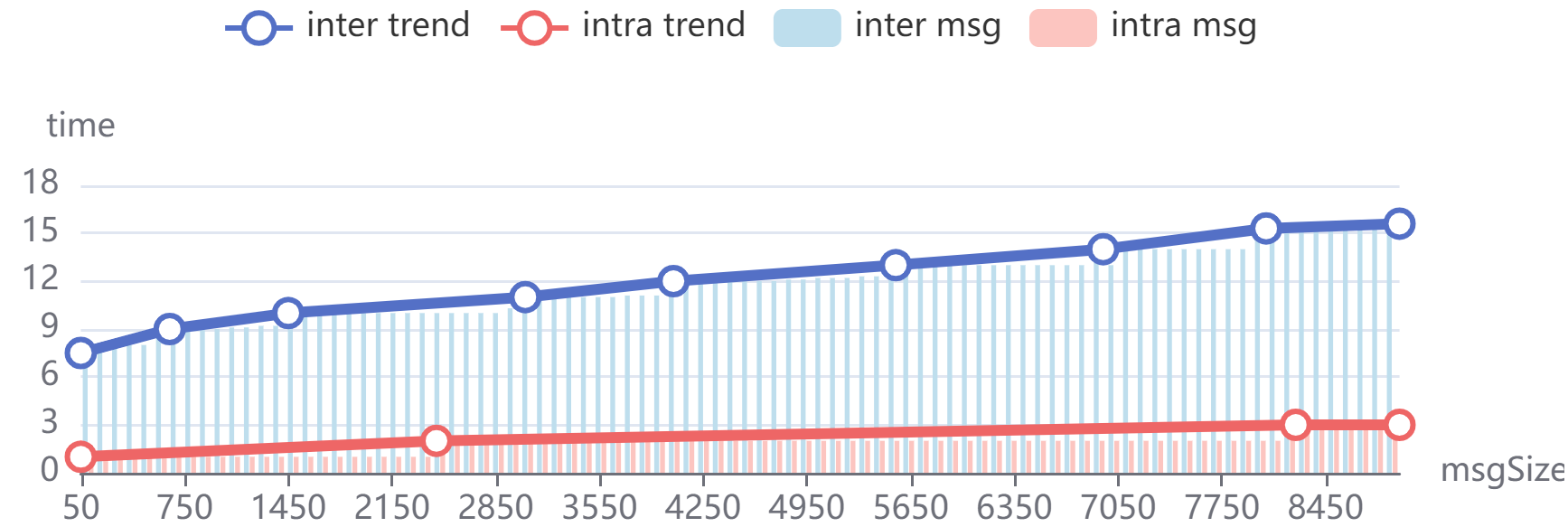}
\caption{Latency criteria of PCL events. The x-axis represents message size, whereas the y-axis represents time. The blue and red lines denote the latency criteria of intra-node and inter-node messages, respectively. The message size is sampled with an interval of 50 bytes.}
\label{fig:criteria}
\end{figure}

First, we separately sample intra-node and inter-node messages and then collect as many as 10,000 messages for each sampling message-size. Second, we sort the messages with the same size in ascending order.  The \textbf{\emph{Latency criteria}} $C$ is defined as the median transmission time for a particular message-size. The latency criteria defined by our method are shown in Figure~\ref{fig:criteria}. As can be seen in the figure, the message transmission time linearly increases along with message size, and the inter-node latency criteria are much higher than the intra-node criteria.

Utilizing these latency criteria, the communication latency of a message can be calculated across three perspectives: spatial latency, temporal evolution, and PCL dependency, employing Equation~\ref{eq:latency}.
\begin{equation}\label{eq:latency}
    L_{msg}=\frac{t_{msg}}{C_{msg}}
\end{equation}
Where $t_{msg}$ is the real transmission time of $msg$, and  $C_{msg}$ is the latency criterion of $msg$. If $L_{msg}>1$, then $msg$ is considered delayed. A larger $L_{msg}$ corresponds to a longer delay.
A high PCL event is defined as the communication event with the larger $L_{msg}$.

The latency of a communication region can be calculated as
\begin{equation}\label{eq:region-latency}
    RL=\frac{1}{n}\sum\nolimits_{i=1}^nL_{msg_{i}}
\end{equation}
Where $RL$  means Region Latency and $n$ is the number of messages contained in the current communication region.

\subsection{Process-Correlation-Tree-Based Spatial Clustering}\label{ssec:spatial-clustering}

This section describes the process-correlation-tree-based spatial clustering method and presents the visualization of communication regions extracted by our clustering method and the communication latency regions extracted by our own latency criteria (Section~\ref{sssec:vis-region})(\textbf{R1}).

\subsubsection{Clustering Method}

To generate communication regions effectively, we introduce a distance metric based on a novel correlation termed \textbf{\emph{Process-Correlation}}. This metric is central to our clustering method and captures the intensity of communication among processes.

\textbf{\emph{Process Communication Structure}}

\begin{figure}[tb]
\centering
\includegraphics[width=8cm]{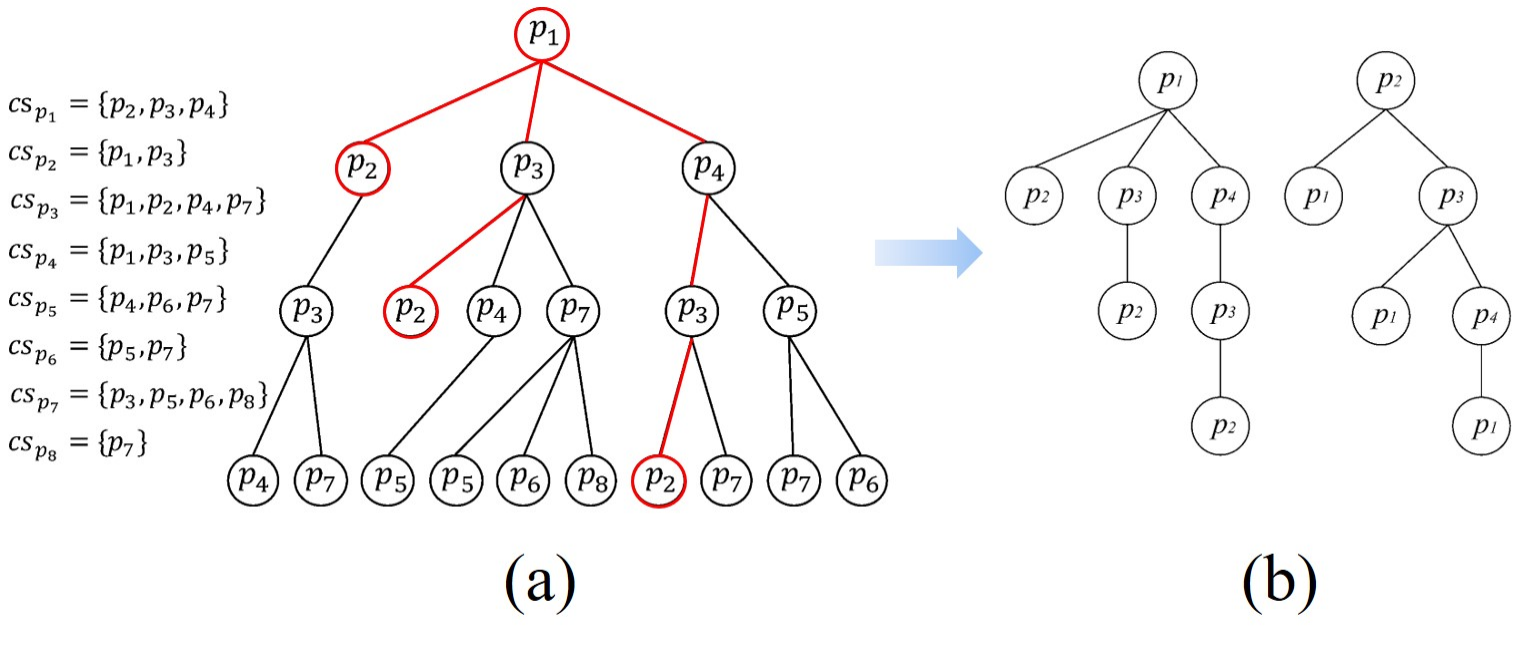} \\
\caption{
The symmetry of process correlations in the process-correlation tree. (a) Process-correlation tree. The dataset on the left describes the process of communication. $cs_{p_{i}}$ denotes the collection of processes communicating with $p_{i}$. The tree on the right describes the correlation of $cs_{p_{1}}$ with the other processes. The red edge represents the path connecting $cs_{p_{1}}$ and $cs_{p_{2}}$.
(b) Process-correlation tree describing the correlation between $cs_{p_{1}}$ and $cs_{p_{2}}$.
}
\label{fig:correlation-tree}
\end{figure}

The process-correlation quantitatively reflects the communication closeness between any pair of processes. For each process $p$, we define $cs_p$ as the set of processes that directly communicate with $p$, either sending or receiving messages. To capture these interactions structurally, we construct a unique communication tree $T_p$ for each process $p$:

\begin{itemize}
\item The root node of $T_p$ is process $p$ itself.
\item The immediate child nodes of $p$ are all processes in the set $cs_p$.
\item Each child node recursively generates its children following the same rule, ensuring uniqueness by the condition that no ancestor and descendant nodes can share the same process ID.
\end{itemize}

Figure~\ref{fig:correlation-tree} provides an illustrative example of the process-correlation tree between process $p_1$ and $p_2$, illustrating the construction process.

\textbf{\emph{Free-Energy Based Distance}}

\begin{algorithm}[!htb]
\caption{Free-Energy Distance Computation} 
\label{alg:free-energy} 
\KwIn{Tree structure $T_p$ for each process $p$}
\KwOut{Distance metric $D(p,q)$}

\tcp{Compute raw process-correlation metric}
\ForEach{process pair $(p,q)$}{
  $R(p,q)\leftarrow\sum_{v\in V_p, v.pid=q}\frac{1}{(v.depth)^2}$\;
}

\tcp{Normalize correlation to transition probabilities}
\ForEach{process $p$}{
  \ForEach{process $q$}{
    $w_{pq}\leftarrow \frac{R(p,q)}{\sum_k R(p,k)}$\;
  }
}

\tcp{Compute partition function via path enumeration or matrix inversion}
\ForEach{process pair $(p,q)$}{
  $Z_{pq}\leftarrow\sum_{\gamma\in\Gamma^{\blacktriangleright}_{pq}}\prod_{e\in\gamma}w_e$\;
}

\tcp{Compute free-energy distance}
\ForEach{process pair $(p,q)$}{
  $D(p,q)\leftarrow-\log(Z_{pq})$\;
}

\tcp{Optional symmetrization step}
\ForEach{process pair $(p,q)$}{
  $D^{\mathrm{sym}}(p,q)\leftarrow\tfrac12\left(D(p,q)+D(q,p)\right)$\;
}

\Return{ $D^{\mathrm{sym}}(p,q)$}
\end{algorithm}




We propose a distance metric based on free energy to quantify the correlation of the process, which effectively captures both direct and indirect communication interactions and satisfies the triangle inequality. Algorithm~\ref{alg:free-energy} summarizes the sequential computational steps required to obtain this metric.

First, the raw correlation $R(p,q)$ quantifies communication intensity based on the tree-structure distances between processes:
\begin{equation}
R(p, q) = \sum_{v \in V_p,\, v.\text{pid} = q} \frac{1}{(v.\text{depth})^2}.
\label{eq:raw-correlation}
\end{equation}

We then convert these correlation scores into row-normalized transition probabilities to interpret them probabilistically. 
Next, we compute the partition function $Z_{pq}$, which aggregates the contributions of all possible directed paths from process $p$ to process $q$:
\begin{equation}
Z_{pq} = \sum_{\gamma \in \Gamma^{\blacktriangleright}_{pq}}\prod_{e \in \gamma} w_e,
\label{eq:partition-function-summary}
\end{equation}
where each $w_e$ represents the transition probability defined from the normalized correlation.

Finally, the symmetric free-energy distance between two processes $p$ and $q$ is given by:
\begin{equation}
D^{\mathrm{sym}}(p, q) = -\frac{1}{2} \left( \log Z_{pq} + \log Z_{qp} \right).
\label{eq:final-free-energy}
\end{equation}

The results of the free-energy distance computation for the example illustrated in Figure~\ref{fig:correlation-tree} are summarized in Table~\ref{tab:correlation}.


Kivimäki \textit{et al.}~\cite{KIVIMAKI2014600} introduced the path-based free-energy distance \(D_{\beta}\), which is proven to be a valid metric: non-negative, symmetric, and satisfying the triangle inequality. Our proposed method adopts a similar framework and provides an intuitive justification of the metric validity, which makes it suitable for clustering applications.


\begin{table}[tb]
\caption{Free-Energy-Based Distance}
\begin{tabular}{|l|l|l|l|l|l|l|l|l|}
\hline
   & p1   & p2   & p3   & p4   & p5   & p6   & p7   & p8   \\ \hline
p1 & 0.00 & 1.18 & 1.44 & 1.33 & 2.72 & 3.40 & 2.82 & 3.22 \\ \hline
p2 & 1.18 & 0.00 & 1.30 & 2.58 & 3.39 & 3.26 & 2.69 & 3.08 \\ \hline
p3 & 1.44 & 1.30 & 0.00 & 1.45 & 2.84 & 2.71 & 1.56 & 2.53 \\ \hline
p4 & 1.33 & 2.58 & 1.45 & 0.00 & 1.35 & 2.61 & 2.84 & 3.24 \\ \hline
p5 & 2.72 & 3.39 & 2.84 & 1.35 & 0.00 & 1.22 & 1.45 & 2.43 \\ \hline
p6 & 3.40 & 3.26 & 2.71 & 2.61 & 1.22 & 0.00 & 1.32 & 2.30 \\ \hline
p7 & 2.82 & 2.69 & 1.56 & 2.84 & 1.45 & 1.32 & 0.00 & 1.15 \\ \hline
p8 & 3.22 & 3.08 & 2.53 & 3.24 & 2.43 & 2.30 & 1.15 & 0.00 \\ \hline
\end{tabular}
\label{tab:correlation}
\end{table}

\begin{enumerate}[leftmargin=*, label=\textbf{\arabic*.}]
\item \emph{Non-negativity.}  
      By definition, we have $0 < Z_{pq}\leq 1$, thus immediately implying 
      \[
      D(p,q) = -\log Z_{pq}\geq 0,
      \] 
      with equality holding if and only if $p=q$ since $Z_{pp}=1$.

\item \emph{Symmetry.}  
      While the partition function $Z_{pq}$ is generally asymmetric (i.e., $Z_{pq}\neq Z_{qp}$), symmetry is enforced explicitly in the optional symmetrized version:
      \[
        D^{\mathrm{sym}}(p,q)=\frac12\left(D(p,q)+D(q,p)\right),
      \]
      ensuring $D^{\mathrm{sym}}(p,q)=D^{\mathrm{sym}}(q,p)$ for all $p,q$.

\item \emph{Triangle inequality.}  
      Consider any three processes $i,k,j$. Since every first-arrival path from $i$ to $j$ can be decomposed uniquely into a first-arrival path from $i$ to $k$, followed by a first-arrival path from $k$ to $j$, the following multiplicative inequality holds:
      \[
        Z_{ij}\geq Z_{ik}\,Z_{kj}.
      \]  
      Taking $-\log(\cdot)$ on both sides, we obtain the additive inequality:
      \[
        D(i,j)\leq D(i,k)+D(k,j).
      \]  
      For the symmetrized version, averaging the inequalities in both directions similarly yields:
      \[
        D^{\mathrm{sym}}(i,j)\leq D^{\mathrm{sym}}(i,k)+D^{\mathrm{sym}}(k,j).
      \]
      Therefore, the free-energy distance satisfies the triangle inequality.
\end{enumerate}

Thus, $D_{\beta}$ is non‐negative, symmetric, and satisfies the
triangle inequality; it can safely replace the original correlation
measure in our hierarchical clustering workflow.

\textbf{\emph{Clustering Workflow}}

The communication data demonstrate a clear hierarchical structure, making it ideal for hierarchical clustering. We employ an agglomerative hierarchical clustering scheme~\cite{Johnson:1967} that merges processes based on the free-energy distance metric. The clustering process is summarized as follows:

\begin{enumerate}
    \item Initially, each process in the set is treated as an individual cluster, forming the leaves of the hierarchical tree.
    
    \item The pairwise linkage free-energy distance between every pair of clusters is computed. In this study, we use the free-energy-based distance as defined in Algorithm~\ref{alg:free-energy}.
    
    \item The two clusters with the smallest distance are merged to form a new non-leaf node in the hierarchical tree.
    
    \item Steps 2 and 3 are repeated until the distance between any two clusters falls below a pre-determined threshold. In this study, the threshold is set to 2, and clustering stops when only a single communication message exists between two clusters.
\end{enumerate}

Applying this method to the data presented in Figure~\ref{fig:correlation-tree} results in two communication regions: \(Region_1 = \{P_1, P_2, P_3, P_4\}\) and \(Region_2 = \{P_5, P_6, P_7, P_8\}\), where communication within each region is much more frequent than between regions.

\begin{figure}[tb]
\centering
\begin{tabular}{cc}
\includegraphics[width=4cm]{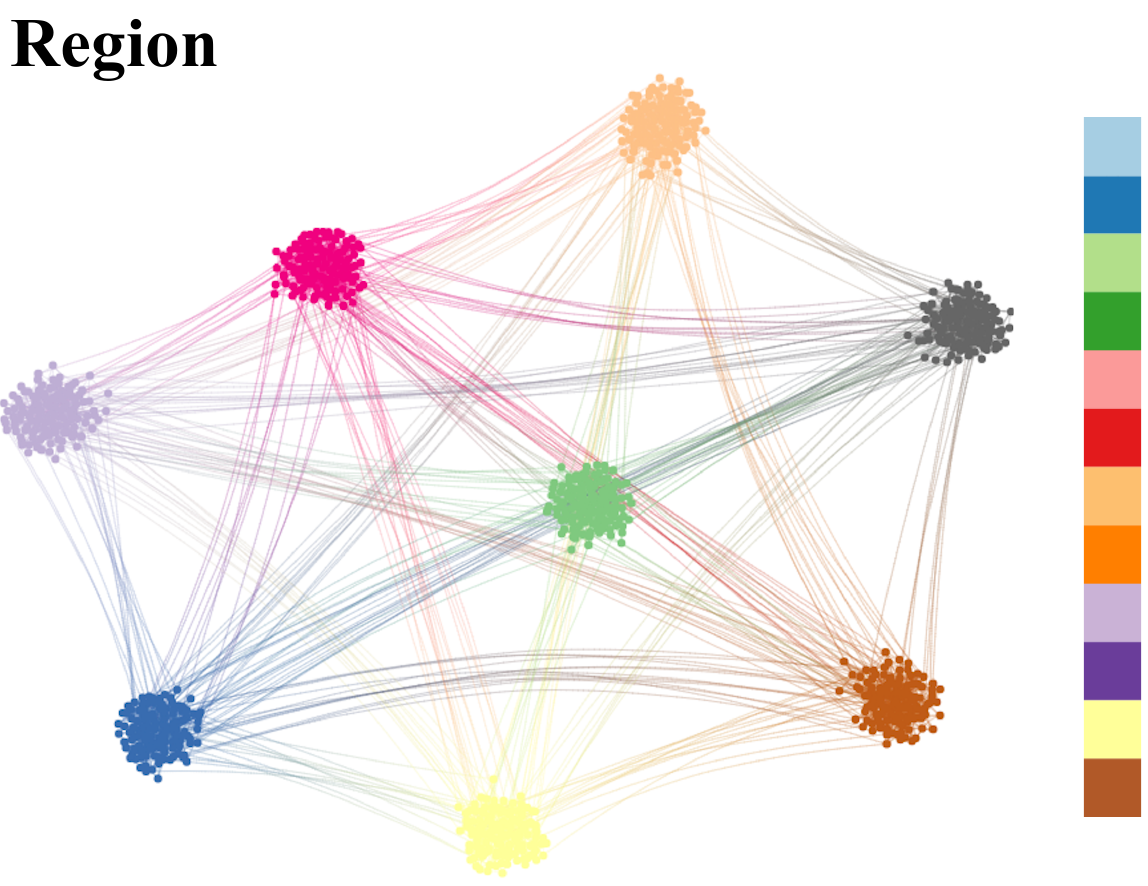}& 
\includegraphics[width=4cm]{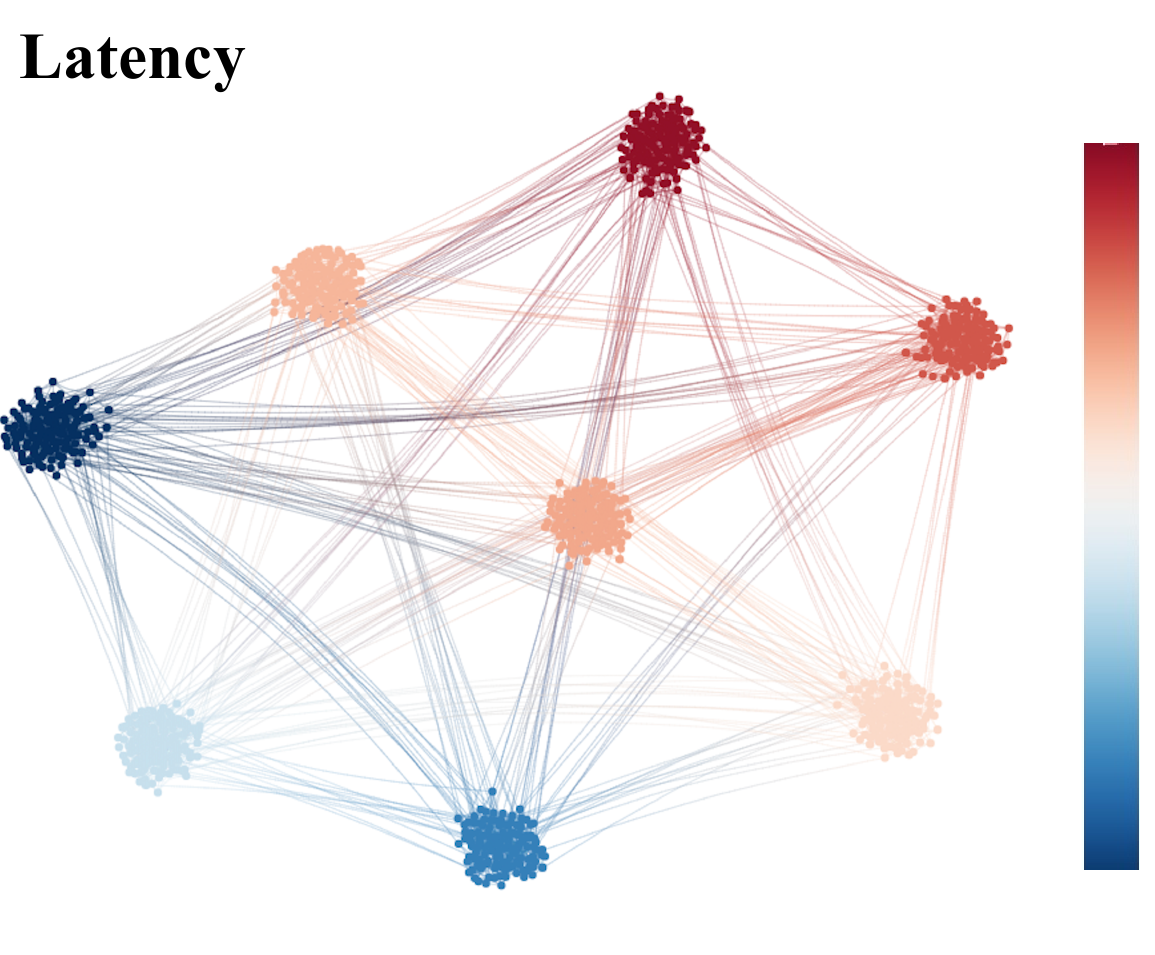}\\
(a) & (b)\\
\end{tabular}
\caption{(a) The communication regions graph, where each communication region is encoded by a certain color. (b) The latency mode of the graph, where the color goes from blue to red, indicates that the latency increases from low to high.}
\label{fig:clustering}
\end{figure}

\subsubsection{Visualization of Communication Regions}\label{sssec:vis-region}
A communication region graph is designed to illustrate the relationship between processes within a region and the relationship between regions.
Figure~\ref{fig:clustering}(a) shows a total of 1024 processes, which are divided into 8 communication regions. A graph node represents a process, whereas a link between two nodes represents the communication between processes.
Each region is represented by a color for users to intuitively understand the complex communication relationship. Our design aims to show the overall clustering results and not just a single process.

Through spatial clustering of process nodes, we find the process nodes with the closest communication and divide them into the same area to further visualize them.
Users can select specific process node regions for detailed analysis according to the clustering results in the communication region graph. 

To improve the visual experience of users, we use 2D force-directed edge building (FDEB), an edge bundling algorithm, to avoid cluttered connections.
The system also provides many functions, including zooming I/O, translating, rotating, and dragging, for users to further observe and analyze the clustering results. 

We also calculate the communication latency of each region using the message latency calculation method provided by the Equation~\ref{eq:region-latency}.
The communication latency of regions is characterized by the communication region graph. As shown in Figure~\ref{fig:clustering}(b), as the color goes from blue to red, the communication latency of a region goes from low to high. This graph synchronously displays the process communication delay information between different regions in real-time. Users can select those regions 
with high communication latency for analysis as shown in Figure~\ref{fig:clustering}.

\begin{figure}[tb]
\centering
\includegraphics[width=3.0in]{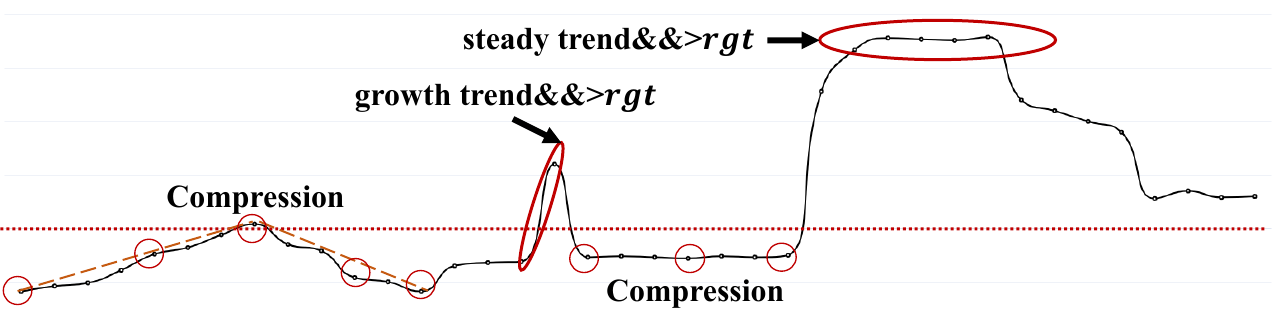}
\caption{Features of communication latency over time, which are used to generate temporal latency abstraction using the sliding window algorithm. The red line denotes $Ave_{region}$, and the important \emph{growth trend period} and\emph{steady trend period} are circled by ellipses. The other features are compressed (reserve part of the data).}
\label{fig:slding-window}
\end{figure}

\section{COMMUNICATION-DEPENDENCY-BASED DAG}\label{sec:dag}
This section discusses the construction of communication-dependency-based DAG Section~\ref{ssec:dag}
(\textbf{R2}), the temporal evolution of PCL events abstraction Section~\ref{ssec:temporal-evolution}, and the communication dependency extraction method Section~\ref{ssec:communication-dependency}.

\subsection{Temporal Evolution of Process Communication Latency Events Abstraction}\label{ssec:temporal-evolution}
A sliding-window-based algorithm is proposed to extract the features of PCL events within communication regions. This algorithm generates an abstraction of temporal evolution.

First, $Ave_{region}$ is defined as the average latency of all messages in the $region$ (Section~\ref{ssec:criteria})
. Second, we define two important features to be extracted, namely, the \emph{growth trend period} and \emph{steady trend period}, as illustrated in Figure~\ref{fig:slding-window}. For other unimportant periods, we reserve three timestamps to compress the time period, namely, start, mid, and end. Based on the features defined above, we use the sliding-window-based algorithm to generate the temporal latency abstraction.

The \emph{growth trend period} refers to a time interval during which the delay data exhibits a sustained increasing trend. This could indicate changes within the communication network leading to a gradual rise in message delays. An example of this feature is depicted in Figure~\ref{fig:slding-window}, showing a steady upward trend in delay data.

Conversely, the \emph{steady trend period} implies that despite stable communication latency, delays are consistently maintained at a higher level, making it difficult to revert to lower values. The illustration in Figure~\ref{fig:slding-window} further clarifies this point, emphasizing fluctuations in message delays at a higher level.

For time intervals of lesser importance, we opt to retain three timestamps denoting the start, midpoint, and end, allowing us to compress these periods effectively.

Building upon the definitions provided above, we employ a sliding-window-based algorithm to generate the abstraction of temporal latency. We believe that this approach better captures the features of events within the communication area, offering valuable insights for subsequent analyses.

\begin{figure}[tb]
\centering
\includegraphics[width=3.5in]{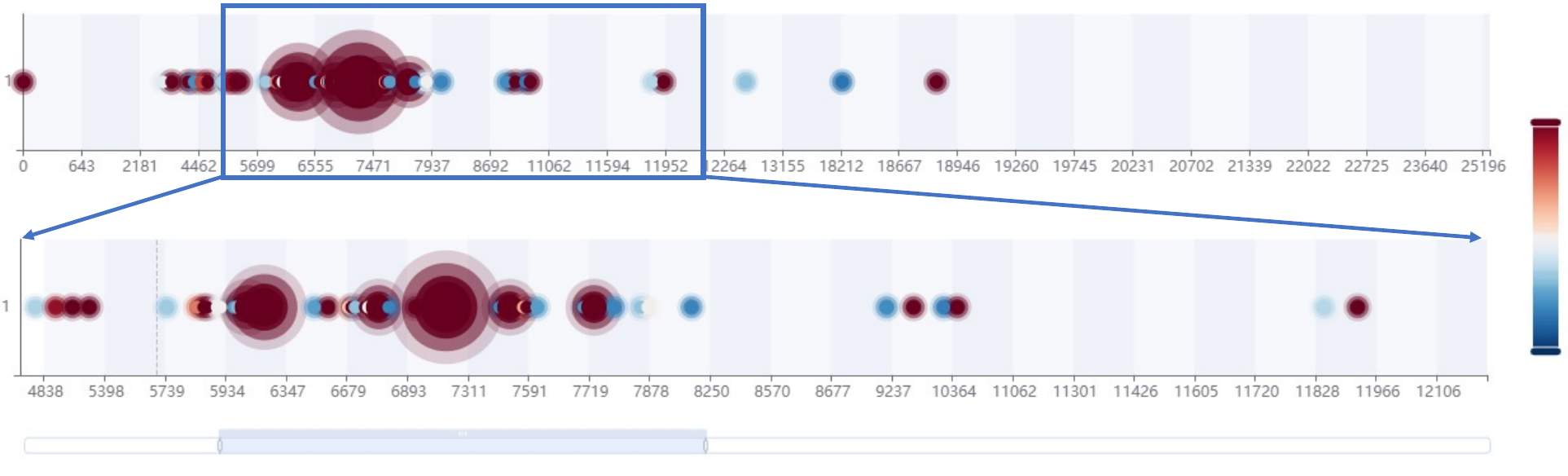}
\caption{Evolution of the latency of a communication region. The abstraction of the communication latency of segments is generated based on temporal latency features. A slider component is used to select a certain period to analyze.}
\label{fig:temporal}
\end{figure}

We also design a temporal evolution view of region communication latency to help users locate those periods with high latency. The temporal evolution view of PCL events shows the overall latency abstraction over time, which can help users interactively explore these events. The DAG (Section~\ref{ssec:dag}) is constructed in the period that users have selected through the slider component below. The symbols in Figure~\ref{fig:temporal} are defined as follows:
\begin{itemize}
\setlength{\itemsep}{0pt}
\setlength{\parsep}{0pt}
\setlength{\parskip}{0pt}
\item \emph{X-axis}: continuous time periods. 
\item \emph{Circle}: Communication latency of a period.
\item \emph{Color of circle}: Communication latency of a region (Equation~\ref{eq:region-latency}).
\item \emph{Size of circle}: Number of messages delayed in a region.
\item \emph{Arrow point}: Details of the communication latency.
\end{itemize}

\subsection{Communication Dependency Among Processes}\label{ssec:communication-dependency}

Based on the physical time order of communication events, the logical time is calculated using the Vector Clock approach. This approach involves the following mathematical formulation:

Based on the physical time order of communication events, the logical time is calculated using the Vector Clock approach. This approach involves the following mathematical formulation:

For each event \(e\), let \(vec_{p}[e]\) represent the logical time associated with event \(e\) belonging to process \(p\). Depending on the type of event:

\begin{enumerate}[label=\arabic*., wide, labelindent=0pt]
    \item In the case of a Send event from process \(i\) to process \(j\), the logical time is updated as follows:
    \begin{equation}\label{time:send}
    {vec_{i}[i] = vec_{i}[i] + 1}.
    \end{equation}
    After updating, process \(i\) sends its updated vector clock to process \(j\) along with the message.

    \item For a Receive event where process \(i\) receives from process \(j\), the logical time is first updated by incrementing process \(i\)'s own clock:
    \begin{equation}\label{time:receive}
    {vec_{i}[i] = vec_{i}[i] + 1}.
    \end{equation}
     \item Then, the vector clock \(vec_{i}\) is updated for all processes \(p\) in the system to ensure it captures the maximum time stamp from both the sender and receiver:
    \begin{equation}\label{time:update}
    {vec_{i}[p] = \max(vec_{i}[p], vec_{j}[p])}.
    \end{equation}

\end{enumerate}

The proposed discipline captures the communication dependency among processes. This discipline enhances the understanding of how events are interconnected within the context of logical time calculations.

\begin{figure}[tb]
\centering
\includegraphics[width=3.5in]{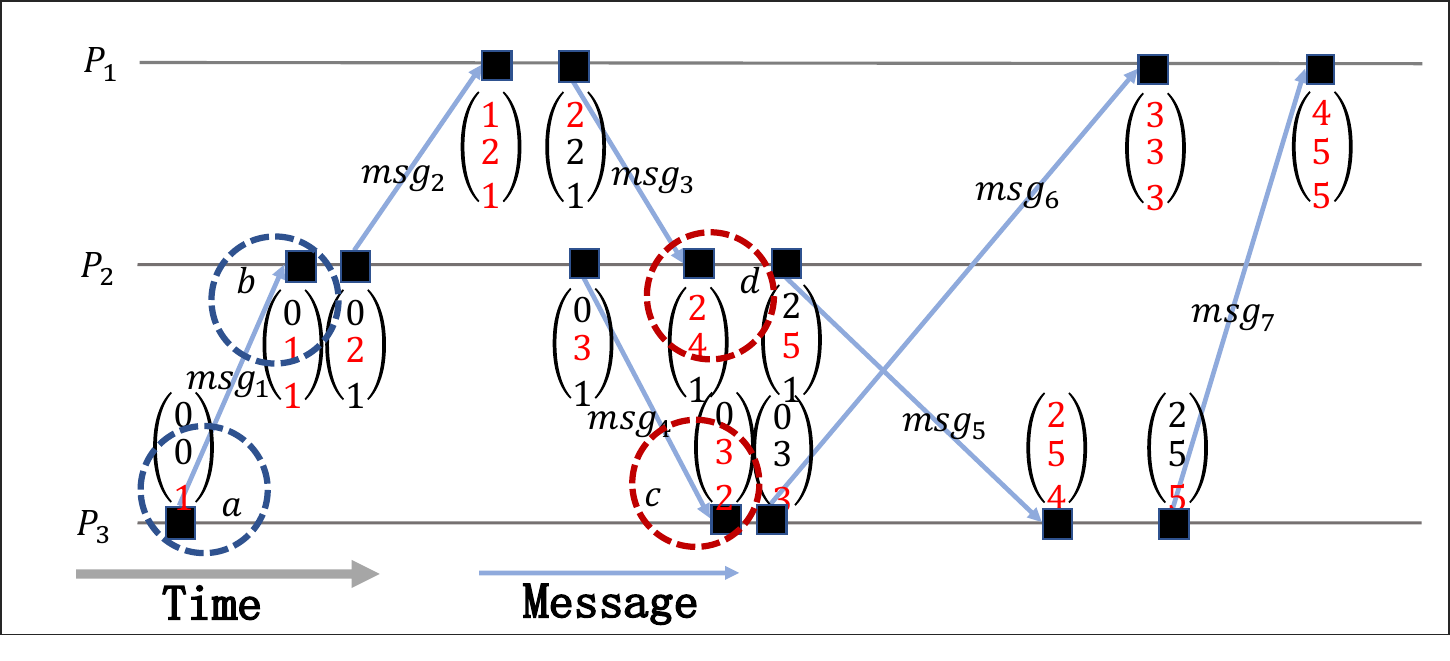}
\caption{Calculating the logical time using Vector Clock. Three processes are involved. The x-axis represents time, whereas the blue line represents the process of communication.}
\label{fig:logical-time}
\end{figure}

We then define the communication dependency among processes based on two disciplines, which are described as follows with reference to Figure~\ref{fig:logical-time}.

\textbf{Discipline 1:} For events $a$ and $b$, $b$ depends on $a$ ($a\to b$). If $\forall i$, $1\leq i\leq N$, then $V_{a}\left [ i \right ]\leq V_{b}\left [ i \right ]$. In Figure~\ref{fig:logical-time}, events $a$ and $b$ are in blue circles. Given that $V_{a}\left [ 1 \right ]=0\leq 0=V_{a}\left [ 1 \right ]$, $V_{a}\left [ 2 \right ]=0\leq 1=V_{a}\left [ 2 \right ]$, $V_{a}\left [ 3 \right ]=1\leq 1=V_{a}\left [ 3 \right ]$, we have $a\to b$ (i.e. $msg1_{send}\to msg1_{receive}$).

\textbf{Discipline 2:} Events $c$ and $d$ are concurrent events that do not depend on each other. If $\exists i$, $1\leq i\leq N$, then $V_{a}\left [ i \right ]> V_{b}\left [ i \right ]$. Figure~\ref{fig:logical-time}, events $c$ and $d$ are in red circles. $V_{c}\left [ 2 \right ]=3\leq 4=V_{d}\left [ 2 \right ]$ and $V_{c}\left [ 3 \right ]=2> 1=V_{d}\left [ 3 \right ]$, $c$ and $d$ are concurrent events. 

Based on the extracted communication dependency, all communication events are ordered by logical time. A logical time-based communication events collection is performed to construct DAG in Section~\ref{sssec:dag}.

\subsection{DAG of Communication State}\label{ssec:dag}
The DAG is constructed using our proposed algorithm (Algorithm~\ref{agl:dag2} in Section~\ref{sssec:dag})
based on the communication dependency described in Section~\ref{ssec:communication-dependency}.
We introduce here our visual designs of the DAG (Section~\ref{sssec:cs-glyph}) as well as a new glyph called CS-Glyph for characterizing the process communication state, the visual layout, and the interactions.

\subsubsection{DAG Construction}\label{sssec:dag}
Based on the communication dependency of events (their logical time order) described in Section~\ref{ssec:communication-dependency}, we construct the DAG using Algorithm~\ref{agl:dag2}. 

\begin{figure}[b]
\centering
\includegraphics[width=3.5in]{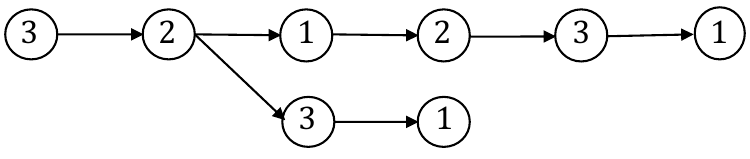}
\caption{The DAG of Figure~\ref{fig:logical-time} calculated by Algorithm~\ref{agl:dag2}. A node represents a process, the number in a node represents process id, and the links represent the messages between processes.}
\label{fig:dependency}
\end{figure}

Figure~\ref{fig:dependency} shows the DAG of communication events presented in Figure~\ref{fig:logical-time}.
A node represents a process, the number in a node represents the process ID, and the links represent the message between processes.
The DAG clearly shows the communication latency among processes.

To further optimize the DAG, we introduce the load balance for processes.
$mc_{p}$ is defined as the sum of messages (sent and received) of process $p$ during a period, whereas $LB_{p}$ represents the load balance of $p$.

First, $AD$ is calculated as
\begin{equation}\label{eq:ad}
    {AD=\frac{\sum_{i=1}^n|mc_{i}-mc_{avg}|}{n}},
\end{equation}
where $AD$ is the arithmetic average deviation, $n$ is the number of processes communicated during this period, and $mc_{avg}$ is equal to $\frac{1}{n}\sum_{i=1}^n mc_{i}$.

We calculate $LB_{p}$ as
\begin{equation}\label{eq:lb}
    {LB_{p}=\frac{|mc_{p}-mc_{avg}|}{AD}}.
\end{equation}

\begin{algorithm}[!htb]
\caption{DAG construction algorithm} 
\label{agl:dag2} 
 \KwIn{the set of communication events $S$ (logical time in ascending order)}
 \KwOut{DAG $G=(V,E)$}
 $process \ V = (pid, eventlist)\leftarrow \emptyset$\;
 $communication \ dependency \ E = edge (V_{i}, V_{j})\leftarrow \emptyset$\;
 
  \ForEach{$ev$ in $S$}{
  \tcp{two types of event: Send and Receive}
    \If{$ev$ is a Send event} { 
        \If{$\exists v\in V, v.pid=ev.pid$}{
            find the newest created node $v_{i}$ (there may exist multiple nodes in $V$ which satisfy the condition) such that $v_{i}.pid=ev.pid$, push $ev$ into $v_{i}.eventlist$\;
        }\Else{
            create a new node $v$, $v.pid=ev.pid$\;
        }
    }
    \BlankLine
    \Else{
        \tcp{$ev$ is a Receive event}
        \If{$\exists v_{i}\in V$, $v_{i}.pid=ev.pid$, $v_{i}.eventlist$ does not contain Send events}{
            push $ev$ into $v_{i}.eventlist$\;
        }
        \Else{
            create a new node $v$, $v.pid=ev.pid$;
        }
        \If{the Send event corresponding to $ev$ is in $v_{j}.eventlist$}{
            create edge $(V_{i}, V_{j})$\ adding into the $E$; 
            
            \tcp{edge $V_{i}\rightarrow  V_{j}$}
        }
    }
  }
\end{algorithm}

\subsubsection{Visual Designs in DAG}\label{sssec:cs-glyph}
We designed CS-Glyph to show the communication state of processes in DAG. This glyph represents a node in DAG.
As shown in Figure~{\ref{fig:cs-glyph}}, the circular CS-Glyph consists of two parts, namely, an ellipse in the center and multiple bar charts in the border. The symbols in the CS-Glyph are defined as follows:

\begin{itemize}
\setlength{\itemsep}{0pt}
\setlength{\parsep}{0pt}
\setlength{\parskip}{0pt}
\item \emph{Ellipse in the center:} a process with an ID number. 
\item \emph{Color of ellipse:} latency of compute node of a process.
\item \emph{Flatness of ellipse:} load balance of a process. 
\item \emph{Greater flatness:} greater load imbalance of a process.
\item \emph{Bar charts inside the border:} messages received by a process.
\item \emph{Bar charts outside the border:} messages sent by a process.
\item \emph{Length of bar chart:} message transmission time.
\item \emph{Height of bar chart:} message size.
\item \emph{Color of bar chart:} latency of the process computed using Equation~\ref{eq:latency}.
\end{itemize}

{\color{red}
\begin{figure}[tb]
\centering
\includegraphics[width=3.4in]{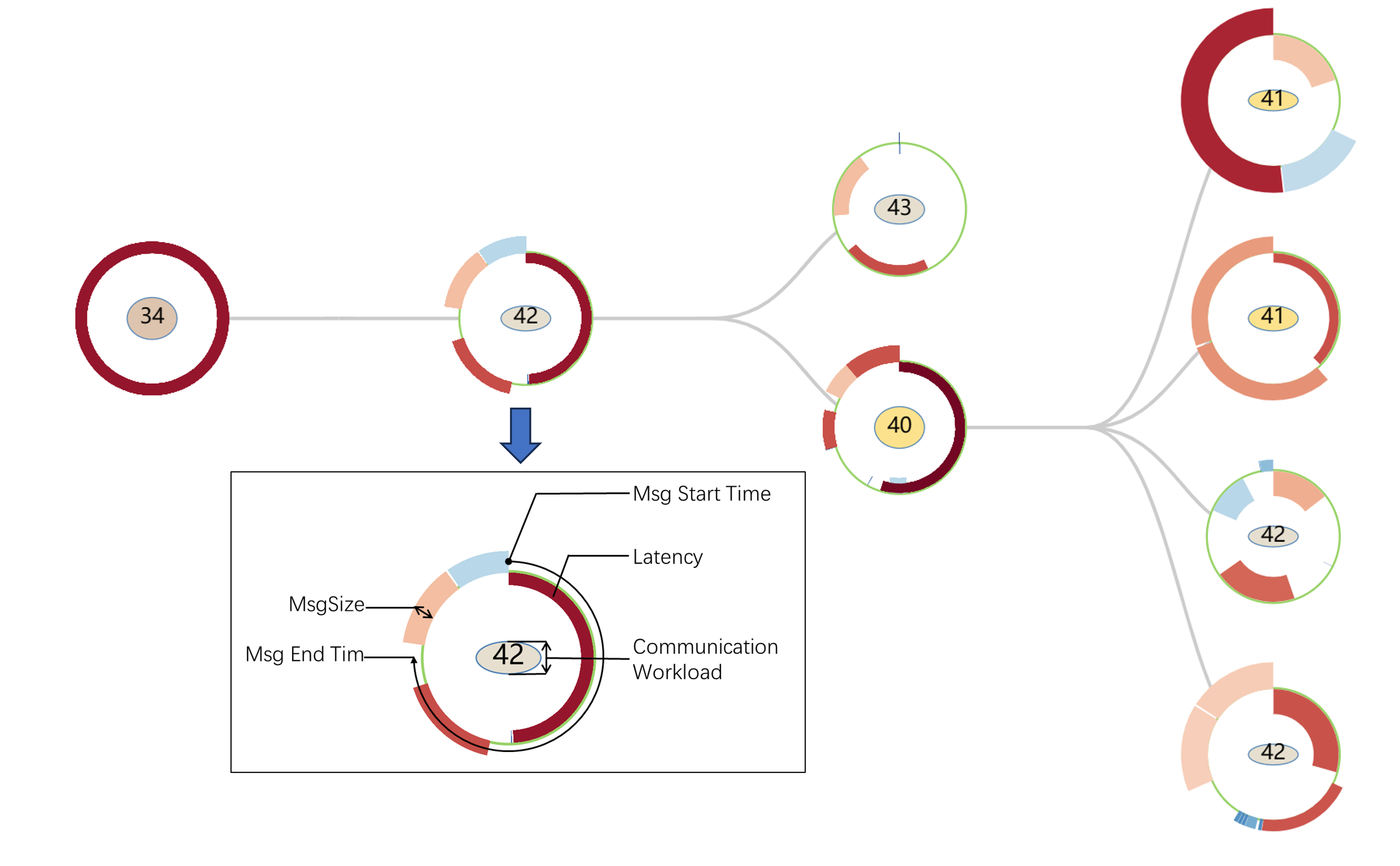}
\caption{Visual design of the CS-Glyph and layout of the DAG. The CS-Glyph is a circle with two parts of information, namely, an ellipse in the center and bar charts in the border. The ellipse represents the process with a process ID, and the bar charts represent the messages sent (outside the border) and received (inside the border).}
\label{fig:cs-glyph}
\end{figure}
}

\begin{figure}[tb]
\centering
\includegraphics[width=3.5in]{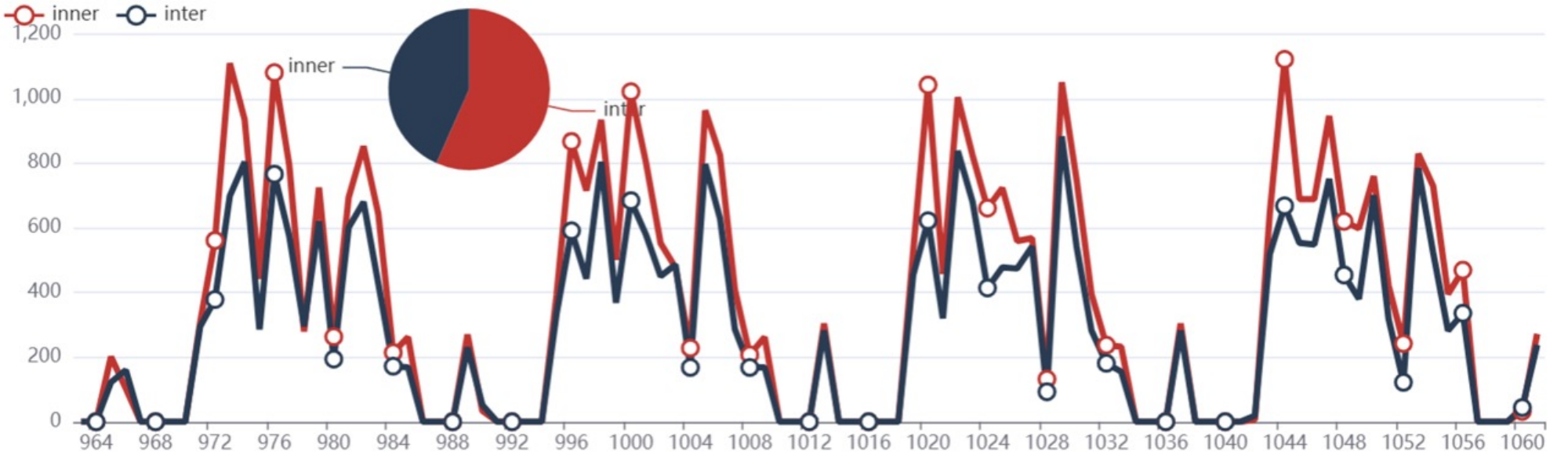}\\
(a)\\
\hfil
\includegraphics[width=3.5in]{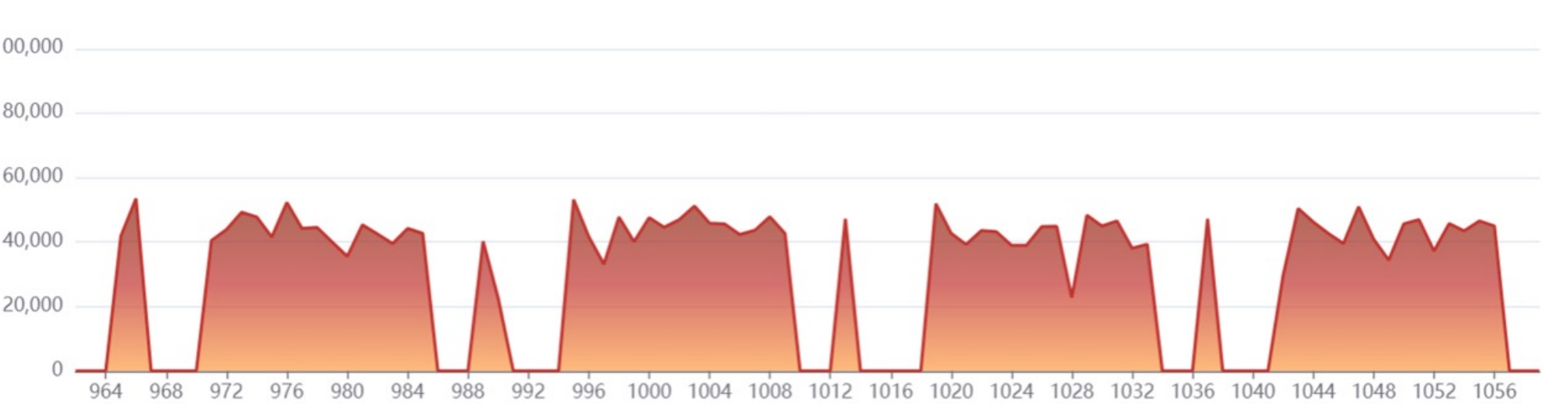}\\
(b)\\
\hfil
\includegraphics[width=3.5in]{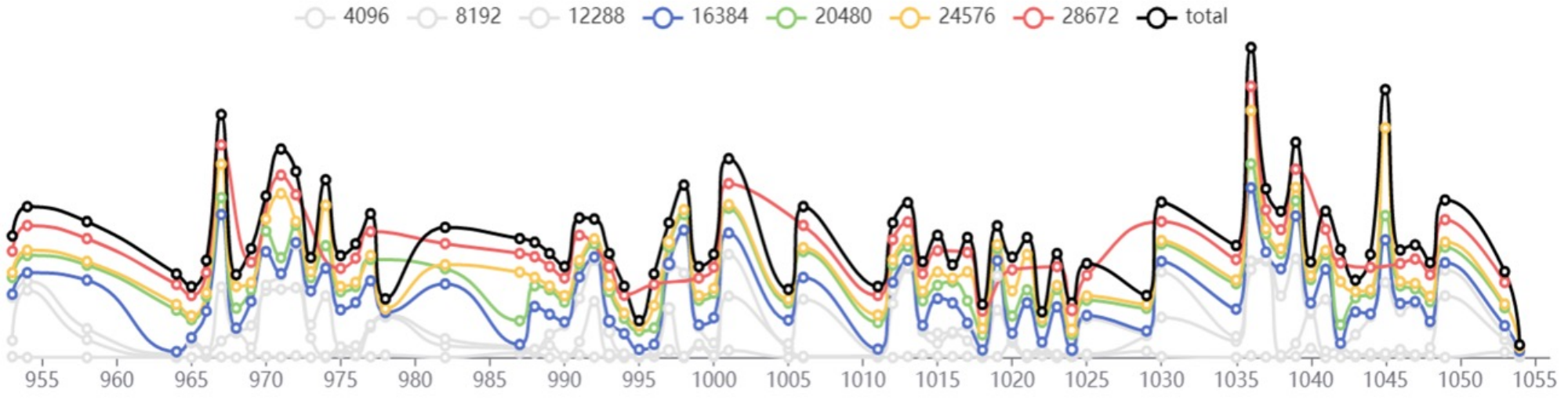}\\
(c)\\
\caption{(a) Mapping analysis view of a period. (b) Communication pattern analysis view of a period. (c) Background analysis view of a period.}
\label{fig:attribution}
\end{figure}

A hierarchical layout and edge-bounding of links are applied to DAG for a clearer display. As shown in Figure~\ref{fig:cs-glyph}, the above designs can address the low rendering efficiency of large-scale process communication.
In the vertical direction, the nodes of DAG are placed on different layers. In the horizontal direction, the nodes show a dependency relationship. 

Some interactions are also offered to help users easily obtain the process communication state, hence facilitating their exploration of communication latency among processes.
The interaction operation is discussed in detail as follows.

\begin{enumerate}
    \item \textbf{Overview and Zooming:} Users can effortlessly access the communication status information for the entire DAG by employing familiar actions like translating and zooming I/O. These interactions facilitate a holistic understanding of communication patterns and statuses across the system.
    
    \item \textbf{Process-Level Control:} For a more focused perspective, users have the ability to selectively view or conceal the communication status of specific processes. Achieved by simply double-clicking on a graph node, this action effectively folds or unfolds the connected nodes to the right, allowing users to tailor their view to their analytical needs.
    
    \item \textbf{Hover Interaction:} A seamless means of acquiring in-depth insights is through cursor hovering. By hovering over a node, users gain detailed information about the associated process. Simultaneously, this action also highlights the pertinent links, offering a comprehensive visualization of the communication dynamics surrounding that specific node.
\end{enumerate}

These interactions enhance user engagement and empower efficient exploration of the communication status within the DAG, ultimately contributing to a more comprehensive understanding of the underlying data.

\begin{figure*}[tb]
\centering
\includegraphics[width=0.85\textwidth]{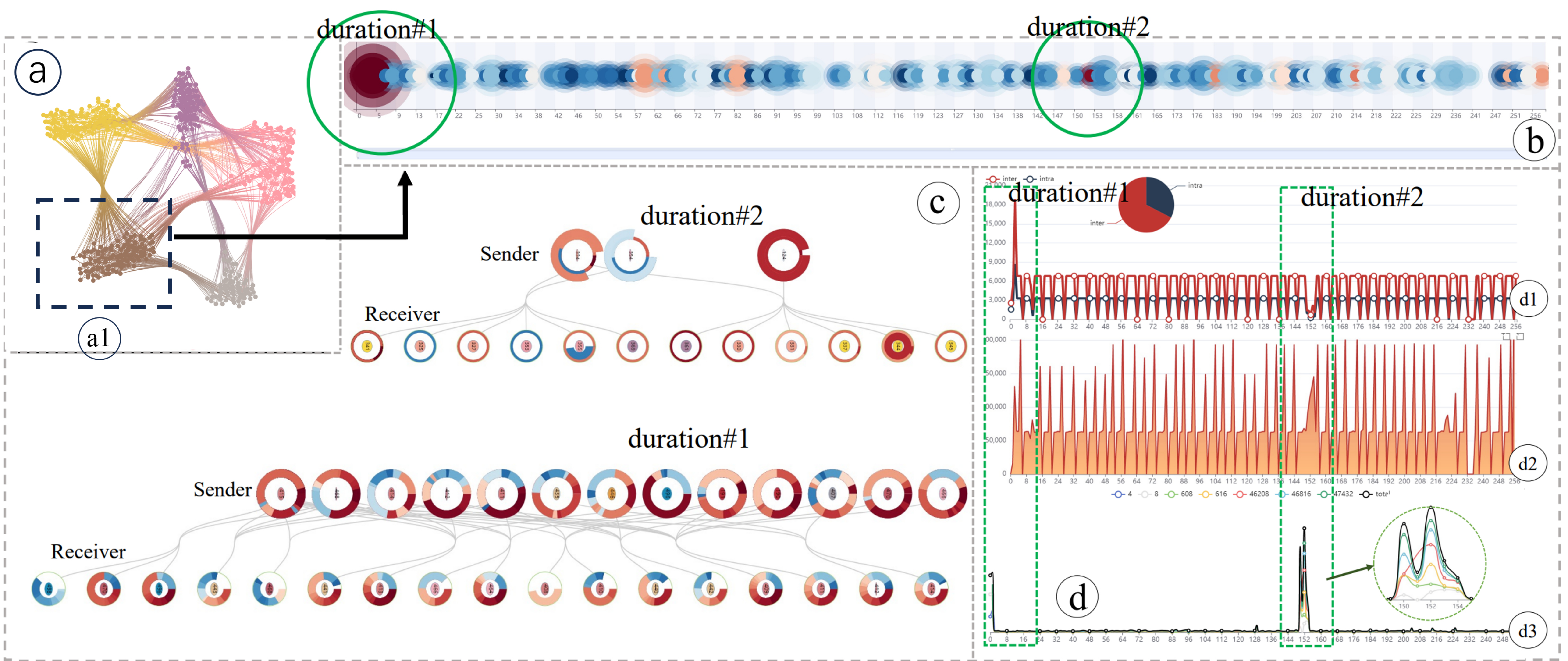}
\caption{Visualization of MiniFE running with 512 processes on TH-1A. (a) is the spatial view showing the process of communication of regions. (b) is the evolution view showing the temporal latency abstraction of a region. (c) is the DAG view showing the communication latency among processes in detail. (d) is the attribution view with three parts, namely, mapping analysis (d1), communication pattern analysis (d2), and background traffic analysis.}
\label{fig:result-1}
\end{figure*}

\begin{figure}[b]
\centering
\includegraphics[width=0.48\textwidth]{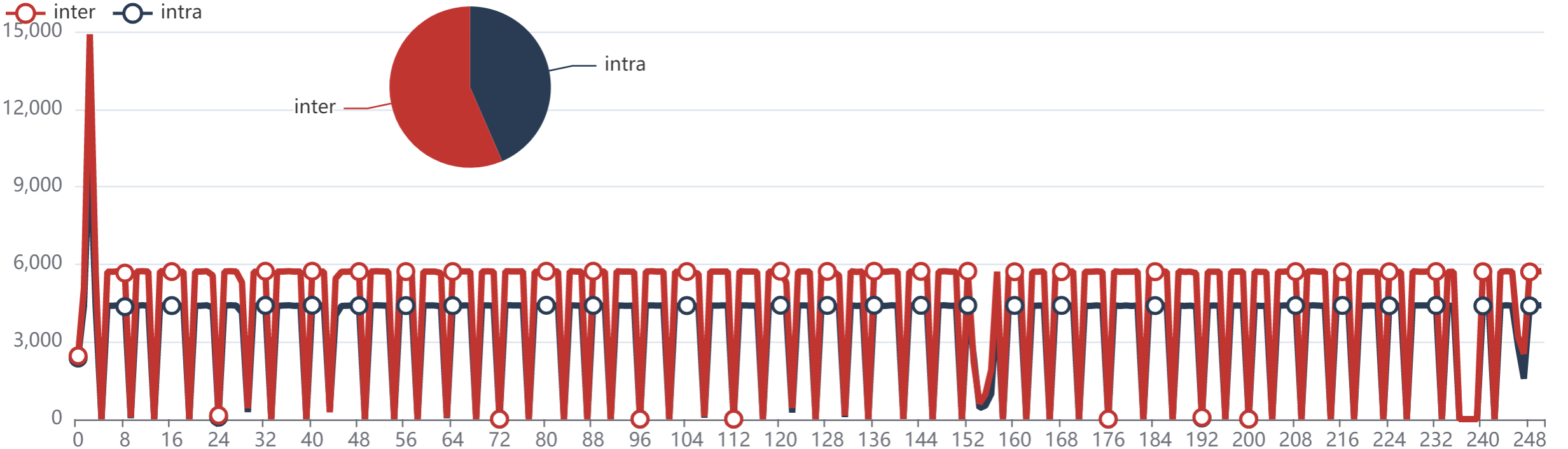}
\caption{Statistics of inter- and intra-node communication after remapping. The number of inter-node messages was reduced from 1385144 to 1159653.}
\label{fig:opt-mapping}
\end{figure}

\begin{figure*}[tb]
\centering
\includegraphics[width=0.85\textwidth]{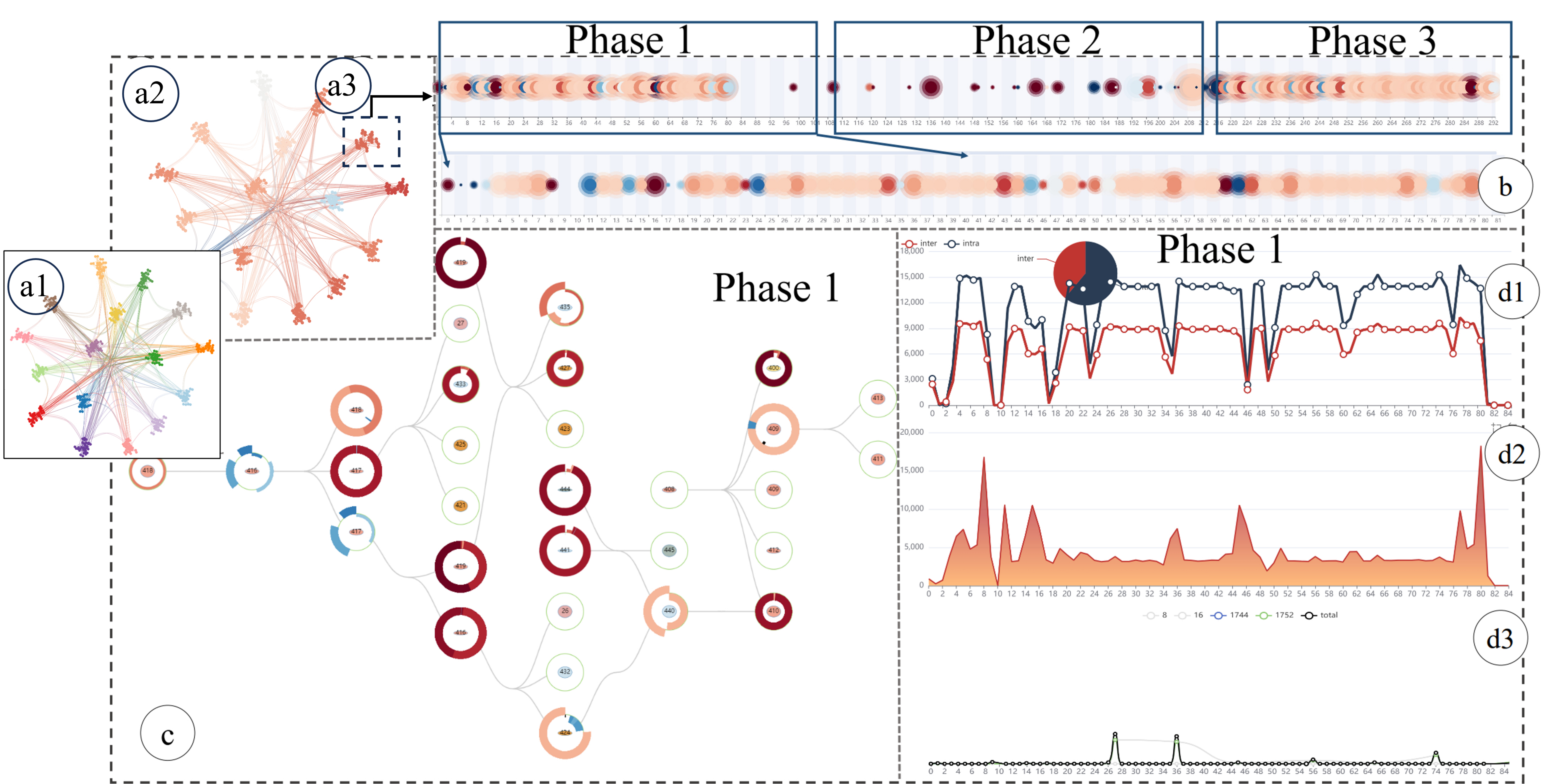}
\caption{Visualization of NPB CG running with 512 processes. (a1) shows the communication latency of regions, where each region is encoded by a certain color. (a2) is the latency mode of (a1). (a3) is a region with high latency. (b) is the latency evolution of region a3, which has three phases. (c) is the DAG graph of \emph{Phase1}, which shows the communication latency among processes. (d) is the attribution view of \emph{Phase1}. Three views are displayed to analyze the causes of PCL events, namely, the mapping (d1), communication pattern (d2), and background traffic (d3).}
\label{fig:result-2}
\end{figure*}

\begin{figure*}[b]
\centering
\includegraphics[width=0.76\textwidth]{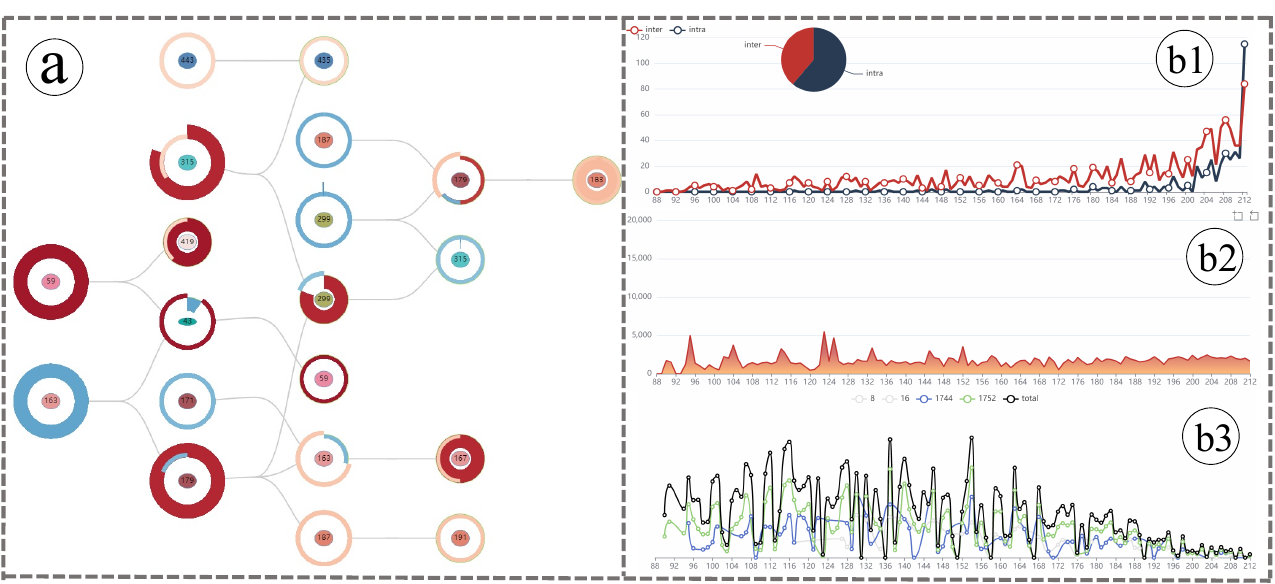}
\caption{Visualization of \emph{Phase2} in Figure~\ref{fig:result-2}. (a) is the DAG graph of \emph{Phase2}. Many CS-Glyphs have dark red bar charts, which indicate high communication latency. Three views are displayed to analyze the causes of PCL events, namely the mapping (b1), communication pattern (b2), and background traffic (b3).}
\label{fig:result-3}
\end{figure*}

\section{PCL Events Attribution Strategy}\label{sec:attribution}

In this section, we analyze the characteristics of different PCL events causes.
Furthermore, we summarize the attribution strategies (\textbf{R3}) and the optimization methods of different PCL events (\textbf{R4}).

\textbf{A1 Poor process-to-processor mappings.}
This problem is caused by too much inter-node communication.
Therefore, poor process-to-processor mapping occurs when DAG has many CS-Glyph pairs with ellipses that are colored differently. 
To address this problem, we present a view that helps users analyze the mapping timing variation of their simulations.
As shown in Figure~\ref{fig:attribution}(a), the mapping view consists of a line chart and a pie chart.
The line chart shows the mapping for each duration, where the x-axis represents the duration, which can be defined by users, whereas the y-axis represents the number of inter- and intra-node communications.
Meanwhile, the pie chart summarizes the mapping as a whole. The red component represents the count of inter-node communications.
A high proportion of the red color denotes poor mapping.

\textbf{O1 Optimize the node mapping scheme.}
To optimize the mapping, users can map the processes that frequently communicate with one another to the same physical node.
Our system prompts users to minimize the communication transmission between physical nodes as much as possible to reduce the hop count of message transmission and the risk of congestion.
The system can also recommend high-efficiency mapping by using the graph partitioning algorithm.

\textbf{A2 Poor communication patterns.}
Generally, communication patterns can describe the process communication load balance. The flatness of
ellipse in the center of CS-Glyph represents this information.
A higher flatness means poor communication patterns.

As shown in Figure~\ref{fig:attribution}(b), we provide an area chart to display the communication load balance of the simulation.
In this view, the x-axis represents the duration, whereas the y-axis represents the average communication load of a process, which we calculate using Equation~\ref{eq:lb}.
The area chart is rendered with a gradient color ranging from yellow to red, with red indicating that the communication load in the current duration is imbalanced.

\textbf{O2 Optimize the simulation communication algorithm. }
Fixing a poor communication pattern requires modifying the communication algorithm of the simulation.
Users can employ combined communication to avoid some poor point-to-point communication patterns, hence improving parallelism and further reducing the traffic for a single process.

\textbf{A3 Background traffic.}
Background traffic varies by time.
If the transmission time of the same size message varies dramatically in the inter-node communication, then we consider this transmission time to be influenced by background traffic.
In the DAG, the bar chart with the same height across all CS-Glyphs has different colors.

The view in Figure~\ref{fig:attribution}(c) records the variance in the transmission time of different messages and reflects the influence of background traffic.
In this view, the x-axis represents the duration, whereas the y-axis represents the average transmission time of different message sizes for each duration.
We calculate and then normalize the average message transmission time. We also color the line as gray where the y value shows the least fluctuation.
A greatly fluctuating y value indicates serious background traffic.

\textbf{O3 Optimize the simulation running environment.}
When performance degradation is caused by background traffic, users do not need to modify the algorithm or communication mapping.
Our system will prompt these users to just wait for the congestion to ease and run the simulation again.

\section{EXPERIMENTAL RESULTS}\label{sec:results}
We analyze in this section the causes of PCL events in two simulations.
We invite three experts to evaluate the system and provide detailed feedback.

\subsection{MiniFE}\label{ssec:minife}
MiniFE is a proxy application for unstructured implicit finite element codes. We run this application with 512 processes on TH-1A. However, this program has poor performance.
As shown in Figure~\ref{fig:result-1}, we use PCLVis to analyze the PCL events in MiniFE to help users improve the communication performance of this parallel program.

Figure~\ref{fig:result-1}(a) shows the 5 communication regions of MiniFE. The processes within each of these regions communicate with one another intensively. We select the yellow region from Figure~\ref{fig:result-1}(a1), which has many PCL events. Figure~\ref{fig:result-1}(b)  shows the PCL temporal evolution of the yellow region. The color of durations 1 and 2 is unusually red, which confirms the presence of communication latency in these durations. Our system only has one DAG for users at the same time. Nevertheless, to better display the effect and save space, Figure~\ref{fig:result-1}(c) flips the DAG for display and places the DAGs of duration 1 and duration 2 together.

\textbf{Analysis of \emph{Duration 1}}: 

To analyze the causes of PCL events in duration 1, we display the communication details of the process inside the yellow region during duration 1 in Figure~\ref{fig:result-1}(c-duration 1).
The upper CS-Glyphs represent the sender, whereas the lower CS-Glyphs represent the receiver.
The bar chart in each CS-Glyph is colored red, thereby indicating that a significant amount of time has been spent on communication.
These CS-Glyphs have the same characteristics. Specifically, the ellipses of CS-Glyph pairs are flat and have different colors, which suggests that those processes belonging to different nodes communicate with one another intensively. 
We combine with Figure~\ref{fig:result-1}(d) for in-depth analysis.
In Figure~\ref{fig:result-1}(d1), the proportion of inter-node communication is larger than that of intra-node communication, and the number of internode communications in duration 1 rapidly increases.
The message transmission time is affected by background traffic when the messages need to be transmitted across inter-node links.
Figure~\ref{fig:result-1}(d3) shows that the transmission time increases dramatically in duration 1, whereas Figure~\ref{fig:result-1}(d2) shows that the communication load balance of MiniFE has no significant impact on performance. Therefore, the PCL events in duration 1 are caused by poor process-to-processor mapping.

Apart from attribution, PCLVis can also provide some possible optimization methods. To address the poor process-to-processor mapping, we propose a remapping algorithm based on graph partition and then remap MiniFE to obtain a better mapping and to reduce the 225491 messages being transmitted across inter-node links (Figure~\ref{fig:opt-mapping}). 

\textbf{Analysis of \emph{Duration 2}}: 

Figure~\ref{fig:result-1}(c-duration 2) shows the communication dependency and state of PCL events in duration 2.
Compared with Figure~\ref{fig:result-1}(c-duration 1), this view has fewer links, and the height of the bar charts is higher.
Figure~\ref{fig:result-1}(d) shows that the number of communication decreases (Figure~\ref{fig:result-1}(d1)), but the transmission time of large messages greatly increases to 47462 bytes (Figure~\ref{fig:result-1}(d3)), hence suggesting that the transmission of large bytes in MiniFE leads to communication latency.
Therefore, the PCL events in duration 2 are caused by background traffic.
When the program is disturbed by background traffic, the system will prompt users to change the job node or adjust the simulation running time.

\subsection{NAS Parallel Benchmarks Conjugate Gradient}\label{ssec:cg}
We then analyze the PCL of \textbf{NAS parallel benchmarks conjugate gradient (NPB CG)} using PCLVis.
These benchmarks are derived from computational fluid dynamics (CFD) applications and consist of five kernels.
We use the conjugate gradient as our kernel, which we run with 512 processes on TH-1A. We easily identify the process mapping problem using PCLVis.
To clearly illustrate other causes of latency, we optimize the mapping using a graph-based mapping algorithm as a pre-processing step.
Therefore, this section will not discuss the mapping problem in detail.

The visualization result is shown in Figure~\ref{fig:result-2}.
As shown in Figure~\ref{fig:result-2}(a1), the 512 processes are divided into 16 communication regions. The communication among processes within each region is similar to that in others. We then switch to the latency mode (Figure~\ref{fig:result-2}(a2)),  and find that the communication latency of one region (Figure~\ref{fig:result-2}(a3)) is higher than that of others. Therefore, we select the specific region with higher communication latency for further analysis.

We initially observe the temporal evolution of communication latency in region a3.Figure~\ref{fig:result-2}(b)  shows 3 communication phases, namely, \emph{Phase1}, \emph{Phase2} and \emph{Phase3}, which we analyze separately as follows.

\textbf{Analysis of \emph{Phase1}}: 

As shown in Figure~\ref{fig:result-2}(c), many CS-Glyphs in the DAG have a dark red color.
To obtain additional information about the communication latency among different processes, we compare these dark red CS-Glyphs with the light-colored CS-Glyphs and find that the ellipses in the former are much flatter than those in the latter, thereby suggesting that the communication load of the dark red processes is very unbalanced.
The DAG also suggests a potential problem with the communication pattern due to the highly unbalanced communication load.

To identify the cause of communication latency, we use the attribution view for further analysis. As shown in Figure~\ref{fig:result-2}(d1), the intra-node communication is much greater than the inter-node communication because we have already improved the process mapping. Meanwhile, Figure~\ref{fig:result-2}(d2) shows that the average load balance of processes greatly varies over time. Multiple peaks can be found in Figure~\ref{fig:result-2}(d2), which indicate a poor communication pattern. The lines in Figure~\ref{fig:result-2}(d3) change smoothly with a low value, which indicates low background traffic in this phase, thereby proving that communication latency is mainly caused by poor communication patterns. 

Therefore, our system prompts the users to modify the simulated communication algorithm to improve the communication pattern.

\textbf{Analysis of \emph{Phase2}}: 

Figure~\ref{fig:result-3}(a) shows many CS-Glyphs in the DAG with a dark red color. We initially check the ellipses in these CS-Glyphs and find that these ellipses have low flatness, thereby confirming that the dark red processes have a balanced communication load.
In other words, this particular phase exhibits a good communication pattern. We also examine the bar charts (representing messages) of the same height (representing message size) and find that messages of the same size have different colors, thereby suggesting that some background traffic may interfere with the communication in our simulation.

To identify the cause of communication latency, we use the attribution view for further analysis.
We observe Figure~\ref{fig:result-3}(b2) directly because the process mapping has already been improved.
We find that the average load balance of processes steadily varies over time and maintains a low value.
This communication pattern can be considered good. 
Furthermore, the lines exhibit significant changes with high values as shown in Figure~\ref{fig:result-3}(b3), thereby confirming a high level of background traffic in this phase and further substantiating that the latency is primarily caused by background traffic.

Our system then prompts the users to just wait for the congestion to ease and run the simulation again.
After analyzing \emph{Phase3}, We find that it is similar to \emph{Phase1}. Therefore, we do not present here the visualization results for \emph{Phase3}.

\section{Discussion}\label{ssec:Expert}

In this section, we have extended invitations to three domain experts to conduct an in-depth analysis of their simulations through the utilization of our PCLVis system. 
The synopsis of their feedback and our subsequent discussions pertaining to the experts' input are presented as follows:

\subsection{Expert feedback on the PCLvis system}

\textbf{The Flowchart of PCLVis System}

Like a traffic jam, all experts want to know “\textbf{where}”, “\textbf{how}”, “\textbf{what}”, and “\textbf{why}”. 
\textbf{Where} is the latency?
\textbf{How} is the latency evolved?
\textbf{What} happened in the processes with latency?
\textbf{Why} does the latency occur?

Our flowchart has fully answered these four questions: the spatial view for locating latency; the evolution view for showing the evolution of the latency; the DAG view for analyzing the 
details; attribution view for the reason of the latency.

\textbf{The Spatial View}

All experts hope to find the worst latency region from their large-scale simulations on supercomputers. Also similar to the traffic jam, the latency would occur in a region, but not one process, because it can propagate from one process to another. 
For locating the latency region, all experts are happy with our spatial view design.
Here is one of their comments:
"It’s so great. The complicated process-communication network with more than 1,000 processes has been instantly visualized into several logic regions. I can also find the worse latency region in the latency mode."

In relation to the clustering algorithms employed within the spatial view, the majority of experts express contentment with methodologies capable of categorizing processes based on their pertinence. Nonetheless, there exists an expert who articulates a desire for heightened temporal efficiency within the current algorithm. This viewpoint seamlessly converges with our forthcoming endeavors and serves as a focal point for our future pursuits.

\textbf{The Evolution View}

The experts also want to know the evolution of one latency region. This is because it can help to trace the source of the latency. 
The experts were satisfied to use the sliding window for finding the period of interest by observing the colored circles, which represent the latency.
Meanwhile, they are also like our temporal latency abstraction strategy, which can successfully hide all unimportant temporal features. By using it, they can quickly locate the most important period for further analysis.

\textbf{The DAG View}

Experts also try to understand the details of the latency among several processes, like their dependencies, load balance, and so on.
They highly rate our DAG view with the following 3 features:
1)	It is a good idea to use DAG for visual analytics of communication dependencies;
2)	Interactive exploration is a good choice for resolving the scalability problem;
3)	The CS-Glyph has been well designed to show enough information for users, including load balance, send/receive messages, latency, and so on.

\textbf{The Attribution View}

Attribution is the most important one for users. All experts found the reasons for the latency in their large-scale simulations by using our PCLVis system.

\begin{enumerate}
    \item[$\bullet$] \textbf{Expert A}: from the result in Figure~\ref{fig:teaser}, I found that there are so many inter-node process communications in the attribution view ($d1$). This is a poor process-to-processor mapping. With guidance from the system-recommended graph division, I have successfully optimized his simulation. The ratio between intra-node communication and inter-node communication has been improved from $252545/281753$ to $325064/209234$.
    \item[$\bullet$] \textbf{Expert B}: Through the system's communication pattern analysis and DAG view, it is found that a large number of load-unbalance cases exist, which is the main reason for the communication latency of my simulation. This is difficult to detect by using existing tools. By checking and revising the simulation code for the unbalanced process IDs, the efficiency of my simulation has been improved. One possible suggestion is to list all possible breakpoints for users, although it may be out of the scope of this system.
    \item[$\bullet$] \textbf{Expert C}: My situation is that the efficiency of my simulation is unstable. I have checked my code again and again but found nothing. With the help of PCLVis, the reason is the communication latency caused by the fluctuation of background traffic.
\end{enumerate}

\subsection{Experts Feedback on Current System Limitations}

Experts acknowledged the system's effective latency analysis, noting its assistance in resolving communication delays.
However, they foresee greater convenience through real-time communication latency analysis. 
This enhancement will allow for easier use and analysis.

\begin{enumerate}
    \item[$\bullet$] \textbf{Expert}: At present, the PCLvis system can help me analyze and locate the latency region in my simulation. But it'd be even cooler if I could perform real-time analysis and latency region location on my running simulations.
\end{enumerate}

\section{Conclusion}\label{sec:conclusion}
In this paper, we proposed the PCLVis framework to visually analyze the causes of PCL in a large-scale simulation.
First, we developed a spatial PCL event locating method.
Second, we designed a process-correlation-tree-based spatial clustering algorithm to generate communication regions.
Third, we designed a communication-dependency-based DAG that can help users interactively explore the communication latency among processes in a communication region.
We also designed a new glyph called CS-glyph to show the communication state of each process.
Before constructing the DAG, we proposed a sliding-window-based method to generate an abstraction of PCL events over time, which displays the evolution of communication latency in a region.
Finally, we developed a PCL event attribution strategy to help users improve the efficiency of their simulations.
Several simulations running on the supercomputer TH-1A were analyzed using the proposed PCLVis system.
Users greatly improved the efficiency of their simulations by following the recommendations of our PCLVis system.
Furthermore, the temporal efficiency of PCLVis necessitates augmentation, serving as the focal point of our imminent pursuits. 
We aspire to achieve real-time analysis of PCL events.

\section*{Acknowledgments}
This work was funded mainly by the National Natural Science Foundation of China under Grant No. 62172294, the Tianjin Natural Science Foundation under Grant No. 23JCYBJC00440, and the Tianjin Science and Technology Project under Grant No. 24YDTPJC00400.

\section*{Declarations}

\textbf{Availability of data and materials}

The MPI communication trace datasets used in this study were generated from large-scale simulation runs on the supercomputer. During the simulations, we collected raw parallel execution traces using the Tuning Analysis Utility (TAU). These raw traces were subsequently preprocessed to extract relevant communication events, including MPI rank, event type, timestamp, source/destination ranks, and message size. The processed datasets were then used for the spatial latency localization, temporal abstraction, and communication-dependency analysis presented in this paper.

\noindent\textbf{Competing interests}

Not applicable.

\noindent\textbf{Authors' contributions}

 Chongke Bi, Xin Gao, and Baofeng Fu designed and implemented the PCLVis framework and developed the algorithms for spatial latency localization, temporal abstraction, and communication-dependency DAG construction. Yuheng Zhao and Siming Chen analyzed and processed the MPI communication trace data from the simulations. Ying Zhao and Lu Yang contributed to the latency attribution strategy, experimental design, and overall supervision of the research.

\noindent\textbf{Funding}

This work was funded mainly by the National Natural Science Foundation of China under Grant No. 62172294, the Tianjin Natural Science Foundation under Grant No. 23JCYBJC00440, and the Tianjin Science and Technology Project under Grant No. 24YDTPJC00400.

\bibliography{sn-bibliography}

\end{document}